\documentclass{article}

\usepackage{arxiv}

\usepackage[utf8]{inputenc} 
\usepackage[T1]{fontenc}    
\usepackage{hyperref}       
\usepackage{url}            
\usepackage{booktabs}       
\usepackage{amsfonts}       
\usepackage{nicefrac}       
\usepackage{microtype}      
\usepackage{lipsum}		
\usepackage{graphicx}
\usepackage{natbib}
\usepackage{doi}
\usepackage{listings}

\usepackage{hyperref}

\usepackage{multirow}
\usepackage{caption}
\usepackage{subcaption}
\usepackage{float}
\usepackage{tabulary}
\usepackage{makecell}

\title{A Scalable Tool For Analyzing Genomic Variants Of Humans Using Knowledge Graphs And Machine Learning}

\author{ \href{https://orcid.org/0000-0001-9102-0709}{\includegraphics[scale=0.06]{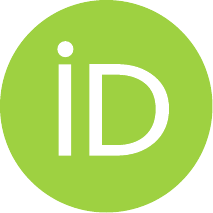}\hspace{1mm}Shivika Prasanna} \\
	University of Missouri-Columbia\\
	Columbia, MA 65201 \\
	\texttt{spn8y@umsystem.edu} \\
	\And
    \hspace{1mm}Ajay Kumar \\
	University of Missouri-Columbia\\
	Columbia, MA 65201 \\
	\texttt{ajay.kumar@missouri.edu} \\
    \And
	\hspace{1mm}Deepthi Rao \\
	University of Missouri-Columbia\\
	Columbia, MA 65201 \\
	\texttt{raods@health.missouri.edu} \\
    \And
    \hspace{1mm}Eduardo Simoes \\
    University of Missouri-Columbia\\
	Columbia, MA 65201 \\
    \texttt{simoese@health.missouri.edu}
    \And 
    \hspace{1mm}Praveen Rao \\
    University of Missouri-Columbia\\
	Columbia, MA 65201 \\
    \texttt{praveen.rao@missouri.edu}
}

\renewcommand{\shorttitle}{\textit{arXiv} Template}

\hypersetup{
pdftitle={A template for the arxiv style},
pdfsubject={q-bio.NC, q-bio.QM},
pdfauthor={Shivika~Prasanna},
pdfkeywords={knowledge graphs, variant-level genomic information, RNA-seq human genome variants, graph machine learning, graph neural networks},
}

\begin{document}
\maketitle

\begin{abstract}
    The integration of knowledge graphs and graph machine learning (GML) in genomic data analysis offers several opportunities for understanding complex genetic relationships, especially at the RNA level. We present a comprehensive approach for leveraging these technologies to analyze genomic variants, specifically in the context of RNA sequencing (RNA-seq) data from COVID-19 patient samples. The proposed method involves extracting variant-level genetic information, annotating the data with additional metadata using SnpEff, and converting the enriched Variant Call Format (VCF) files into Resource Description Framework (RDF) triples. The resulting knowledge graph is further enhanced with patient metadata and stored in a graph database, facilitating efficient querying and indexing. We utilize the Deep Graph Library (DGL) to perform graph machine learning tasks, including node classification with GraphSAGE and Graph Convolutional Networks (GCNs). Our approach demonstrates significant utility using our proposed tool, VariantKG, in three key scenarios: enriching graphs with new VCF data, creating subgraphs based on user-defined features, and conducting graph machine learning for node classification.
\end{abstract}

\keywords{knowledge graphs \and graph machine learning \and graph neural networks \and RNA-seq human genome variants}

\section{Introduction}
\label{introduction}

\cite{olson1993human} aimed to sequence the entire human genome, resulting in an official gene map. The gene map has offered crucial insights into the human blueprint, accelerating the study of human biology and advancements in medical practices. This information has been represented in Variant Calling Format (VCF) files that store small-scale information or genetic variation data. 

Variants are genetic differences between healthy and diseased tissues or between individuals of a population. Analyzing variants can tremendously help prevent, diagnose, and even treat diseases. The process of analyzing these genetic differences or variations in DNA sequences and categorizing their functional significance is called variant analysis. RNA sequencing is similar to DNA sequencing but differs in its extraction. RNA is extracted from a sample and then reverse transcribed to produce what is known as copy or complementary DNA called cDNA. This cDNA is then fragmented and run through a next-gen sequencing system. Examining DNA provides a static picture of what a cell or an organism might do, but measuring RNA tells us precisely what the cell or organism is doing. Another advantage of RNA sequencing is that molecular features sometimes can only be observed at the RNA level.

Variant calling pipeline is the process of identifying variants from sequence data. To measure the deleteriousness of a variant, the Combined Annotation Dependent Depletion (CADD) \cite{rentzsch2019cadd, rentzsch2021cadd} scores tool is used. CADD evaluates or ranks the deleteriousness of a single nucleotide, insertion, and deletion variants in the human genome. The COVID-19 genetic data discussed in this paper was collected from the European Nucleotide Archive (ENA) \cite{ieeedataport}. 

Knowledge graphs such as YAGO \cite{mahdisoltani2013yago3}, Wikidata \cite{vrandevcic2014wikidata}, DBPedia \cite{lehmann2015dbpedia}, and Schema.org \cite{guha2016schema} are crucial for structuring and linking vast amounts of diverse data, to enable efficient information retrieval, enhance data interoperability, and provide a foundation for advanced applications in domains not limited to semantic search, natural language processing, and data integration. One such example is the work of \cite{dong2018challenges} which focuses on constructing a comprehensive knowledge graph (KG) called ProductKG from Amazon's large-scale and diverse product catalog. ProductKG captures product attributes, categories, and relationships using graph mining and embedding techniques. This structured representation aims to improve understanding and retrieval of product information, thereby enhancing user experience and supporting various AI applications within Amazon's ecosystem. Another such example of KGs is FoodKG, introduced by \cite{gharibi2020foodkg} that demonstrates the importance of knowledge graphs in the Food domain through their tool, FoodKG, by integrating diverse datasets, using NLP to extract meaningful entities and state-of-the-art model to enhance and enrich graphs based on Food, Energy and Water (FEW) used by the tool. Their contribution highlights the significant role of knowledge graphs in managing and utilizing large-scale, heterogeneous data.  

Representing genomic data as knowledge graphs allows vast and diverse information from various sources to be integrated. These specialized graph structures, which model entities as nodes and relationships as edges, provide an ideal framework for integrating and organizing diverse biological information from multiple sources. Furthermore, it allows for efficient querying and indexing and supports inference and new knowledge discovery.

The key contributions of this work are:

\begin{itemize}
    \item We introduce a data collection pipeline to extract variant-level genetic information from RNA-seq IDs using the ENA browser. We also discuss adding additional information to the VCF files using the SnpEff tool.
    \item We elaborate on the construction and enrichment of our knowledge graph. We utilize the vcf2rdf tool from Sparqling-genomics for the VCF files, in addition to our explicitly-defined ontologies. We translate the RDF-triples into an efficient and easily parsable format (NQuads). 
    \item We demonstrate the use of our tool, VariantKG and present three scenarios which are \textit{Graph Enrichment} to consume new VCF information or use the existing data from our knowledge base, \textit{Graph Creation} to create subgraphs with user-defined features and \textit{Graph Machine Learning and Inference} for node classification tasks using GraphSAGE and Graph Convolutional Network supported by Deep Graph Library.
\end{itemize}

The remainder of the paper is organized as follows: \ref{related-work} explores the previous works that integrate knowledge graphs with the vast genomic data; \ref{kg-construction} discusses the data collection pipeline, construction of our knowledge graph (KG) using ontology and enrichment of the KG with patient metadata; \ref{graph-storage-database} introduces the graph storage and database used in this research work and tool; \ref{kg-inference} discusses the use of Deep Graph Library (DGL) for node classification tasks and introduces our tool, VariantKG in which we demonstrate 3 scenarios, namely how a user can use data from our existing knowledge base, add new VCFs to enrich the dataset, and use a combination of the added files and the existing knowledge base to perform node classification tasks using two Graph Neural Networks - GraphSAGE and Graph Convolutional Networks (GCNs), using our knowledge graph database for graph machine learning tasks; \ref{conclusion} concludes our work.

\section{Related Work}
\label{related-work}

Knowledge graphs are widely used to integrate and analyze diverse genomic data, providing a comprehensive and contextual representation of genomic information. For instance, \citep{feng2024genomickb} extended the development of GenomicKB by creating a graph database that integrates human genome, epigenome, transcriptome, and 4D nucleome data. This extensive database, annotated from over 30 consortia and portals, includes 347 million entities, 1.36 billion relations, and 3.9 billion properties, covering comprehensive data on pancreases and diabetes' GWAS, disease ontology, and eQTL data. Another notable work, \citep{feng2023genomickb}, presented a knowledge graph, GenomicKB, that consolidates various sources of genomic information, including data from genomic databases, experimental studies, literature, and public repositories, into a single, unified framework. This integration facilitates efficient data analysis and knowledge discovery through a user-friendly web portal.

Knowledge graphs have been instrumental in understanding the COVID-19 virus and its treatment. For example, \citep{sakor2023knowledge4covid19} developed a framework that integrates diverse COVID-19 drug resources to discover drug-drug interactions among COVID-19 treatments, utilizing RDF mapping language and Natural Language Processing (NLP) to extract relevant entities and relationships. Similarly, \citep{reese2021kg} proposed KG-COVID-19, a knowledge graph framework that integrates heterogeneous data on SARS-CoV-2 and related viruses, supporting downstream tasks such as machine learning, hypothesis-based querying, and user interface exploration. \citep{chen2021covid19} used semantic web technology RDF to integrate COVID-19 data extracted from iTextMine, PubTator, and SemRep biological databases into a standardized Knowledge Graph (KG). This COVID-19 KG supports federated queries on the semantic web and is accessible through browsing and searching web interfaces, with a RESTful API for programmatic access and RDF file downloads.

Deep learning has significantly influenced a wide range of domains, with genomic studies being particularly impacted. For instance, \citep{liu2020deepcdr} introduced DeepCDR, a method using deep learning to predict cancer cells' response to different drugs, facilitating effective cancer treatment. Another innovative approach was proposed by \citep{lanchantin2019graph}, who developed ChromeGCN for predicting epigenetic states using sequences and 3D genome data. ChromeGCN leverages graph convolutional networks (GCNs) to predict the epigenetic states of genomic regions, representing genomes as graphs where nodes are genomic regions and edges represent relationships between them. This method's predictive power enables the identification of functional genomic elements and regulatory regions, providing insights into gene regulation and cellular function.

Additionally, \citep{harnoune2021bertclinical} proposed constructing knowledge graphs from clinical data using the BERT (Bidirectional Encoder Representations from Transformers) model. This approach focused on creating biomedical knowledge graphs by leveraging BERT's contextual understanding capabilities to process biomedical text data, including clinical records and scientific literature, extracting meaningful and contextually rich information. \citep{domingo-fernandez2021covid19} developed a multi-modal cause-and-effect COVID-19 knowledge model using Biological Expression Language (BEL) as a triple (i.e., source node-relation-target node) with metadata about nodes. This model utilized GraphML, NDEx, and SIF representations for network visualization and was made accessible through a web platform to enhance its visibility and utility.

Lastly, \citep{al-obeidat2020cone-kg} focused on extracting and utilizing knowledge from COVID-19-related news articles, providing a platform for researchers, data analysts, and data scientists to investigate and recommend strategies to address global challenges. \citep{sun2018knowledge} proposed KGBSVM (Kernelized Generalized Bayesian Rule Mining with Support Vector Machines), a method to analyze high-dimensional genome data, aiming to improve classification accuracy on general tasks, whether binary or multi-class. This method enhances the accuracy and efficiency of genomic data classification, contributing to better data analysis and interpretation in the field.

These diverse efforts highlight the significant potential of integrating knowledge graphs with genomic data and deep learning, facilitating comprehensive data integration, efficient analysis, and innovative solutions to complex problems in genomics and beyond.

\section{Knowledge Graph Construction}
\label{kg-construction}

In this section, we explain how to construct knowledge graphs. First, we describe how the data was collected and the workflow processes. We also incorporate the patient metadata used to enhance the knowledge base. Then, we highlight the process of annotating the data, followed by the ontology description. Lastly, we discuss the representation of VCF files and CADD Scores in a knowledge graph.

\subsection{Data Collection}
\label{data-collection}

\begin{figure}[tbh]
    \centering
    \includegraphics[width=0.8\textwidth]{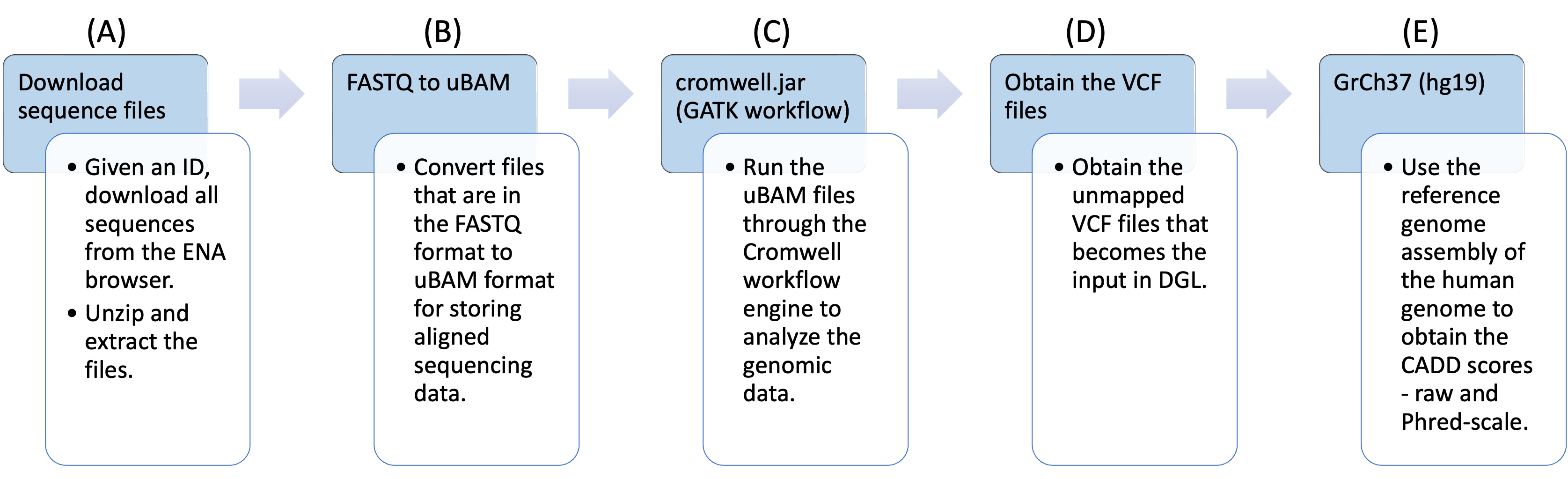}
    \caption{Workflow to demonstrate raw data collection to further process and collate into a dataset.}
    \label{fig:seq-workflow}
\end{figure}

COVID-19 RNA sequence IDs were first collected from the European Nucleotide Archive. A total of 3,716 VCF files were collected till March 2023. The workflow has been shown in Figure \ref{fig:seq-workflow}.

\begin{itemize}
    \item FASTQ files (part A): These IDs were utilized to download the RNA sequences, which were in FASTQ \cite{li2008mapping,li2009fast} format. The FASTQ file format consists of a series of records, each of which contains four lines of text: the first line starting with `@' contains a sequence identifier, the second line contains the actual nucleotide sequence, the third line starts with `+' and may optionally contain additional information about the sequence, and the fourth line contains quality scores encoded as ASCII 10 characters. The quality scores indicate the confidence in the accuracy of each base call and are typically represented as Phred scores.
    \item uBAM files (part B): The FASTQ files were then converted to unmapped BAM (uBAM) files for storing aligned sequencing data. A uBAM file contains unmapped reads, meaning reads that could not be confidently aligned to a reference genome. These reads can be used for downstream analysis, such as de novo assembly, quality control, and identification of novel sequences.
    \item GATK workflow (part C): The uBAM files were passed through the Genomic Analysis Toolkit (GATK) workflow \cite{mckenna2010genome} that converts the files into Variant Calling Format (VCF) \cite{danecek2011variant} files. It is a comprehensive toolkit developed by the Broad Institute that includes various tools and algorithms for processing genomic data, such as read mapping, local realignment, base quality score recalibration, variant calling, and variant filtering. 
    \item The unannotated VCF files that were obtained as the result of the workflow have been shown in part D. For each VCF file, there is also a corresponding CADD Scores file that was obtained using GrCh37 through the workflow, as shown in part E.
\end{itemize}

\subsection{Data Annotations}
\label{data-annotations}

Once the RNA sequencing was completed, two main files were generated for each RNA-seq ID – a VCF file and a CADD Scores file.

For further annotations, SnpEff \cite{cingolani2012program}, a command-line, variant annotation, and effect prediction tool, was utilized. This tool annotates and predicts the effects of genetic variants. SnpEff takes the predicted variants (SNPs, insertions, deletion, and MNPs) as input and produces a file with annotations of the variants and the effects they produce on known genes. 

\begin{figure}[H]
    \centering
    \includegraphics[keepaspectratio, width=\textwidth]{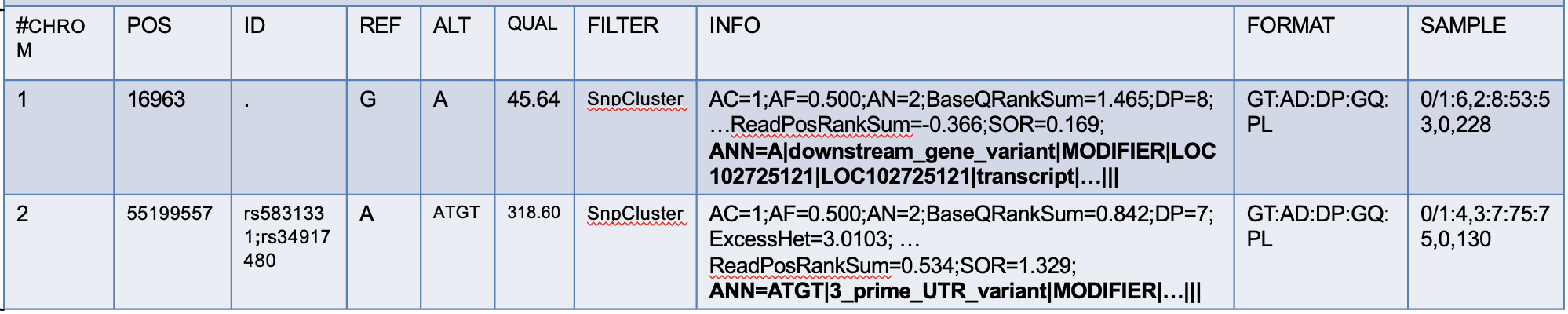}
    \caption{Additional annotations by the SnpEff tool.}
    \label{fig:ann-field}
\end{figure}

SnpEff classifies variants as single nucleotide polymorphisms (SNPs), insertions, deletions, multiple-nucleotide polymorphisms, or an InDel. While the original VCF file contains the INFO field, SnpEff adds additional annotations to this field to describe each variant further. In the process, it also updates the header fields. This field is tagged by `ANN', which is pipe symbol separated and provides a summary of the predicted effects of a genetic variant on each affected transcript. Figure \ref{fig:ann-field} shows the ANN field highlighted in bold.

A variant may have one or more annotations and multiple annotations are comma-separated. There are several fields within the ANN tag, mainly:
\begin{itemize}
    \item Allele (ALT): Information on alternate alleles that are predicted to cause a functional impact on a gene or protein
    \item Annotation (effect): Type of effect caused by the variant on the transcript
    \item Putative impact: Qualitative measure of the impact of the variant on the transcript
    \item Gene name: Name of the affected gene
    \item Gene ID: Unique identifier of the affected gene
\end{itemize}

\subsection{Metadata }
\label{metadata}

The metadata was downloaded from the SRA \footnote{\url{(https://www.ncbi.nlm.nih.gov/sra)}} web tool of NCBI. The metadata runs for each patient are stored in the \textit{Study} section. We leveraged the SRA web tool over entrez to extract a wider range of metadata information such as disease, age, fatality. The obtained metadata was then converted to NQ triples to facilitate the use of named graph to link this information to the variant information obtained, as shown in Section \ref{data-collection}. An example of the metadata obtained for the accessionID SRR12570493 is as follows:

\begin{lstlisting}
SRR12570493,65 (Age),,RNA-Seq,119,1473875214,PRJNA661032,SAMN15967295,582422941,,GEO,Severe COVID,Alive,public,"fastq,sra,run.zq","gs,ncbi,s3","s3.us-east-1,gs.US,ncbi.public",SRX9058173,NextSeq 500,GSM4762164,,PAIRED,cDNA,Homo sapiens,TRANSCRIPTOMIC,ILLUMINA,2021-01-27T00:00:00Z,,2020-09-02T12:06:00Z,1,GSM4762164,male,6,Patient 14 blood,SRP279746,Patient 14,Blood,,,,,,,,,,,,
\end{lstlisting}

The NQ triple for the age attribute for the same accessionID run is as follows:

\begin{lstlisting}
   <https://www.ncbi.nlm.nih.gov/sra/?term=SRR12570589> <https://www.wikidata.org/wiki/Q11904283> "61.0"^^<http://www.w3.org/2001/XMLSchema#float> <sg://SRR12570589> .
\end{lstlisting}

\subsection{Ontology}
\label{ontology}

\begin{figure}[H]
    \centering
    \includegraphics[keepaspectratio, width=\textwidth, height=0.3\textheight]{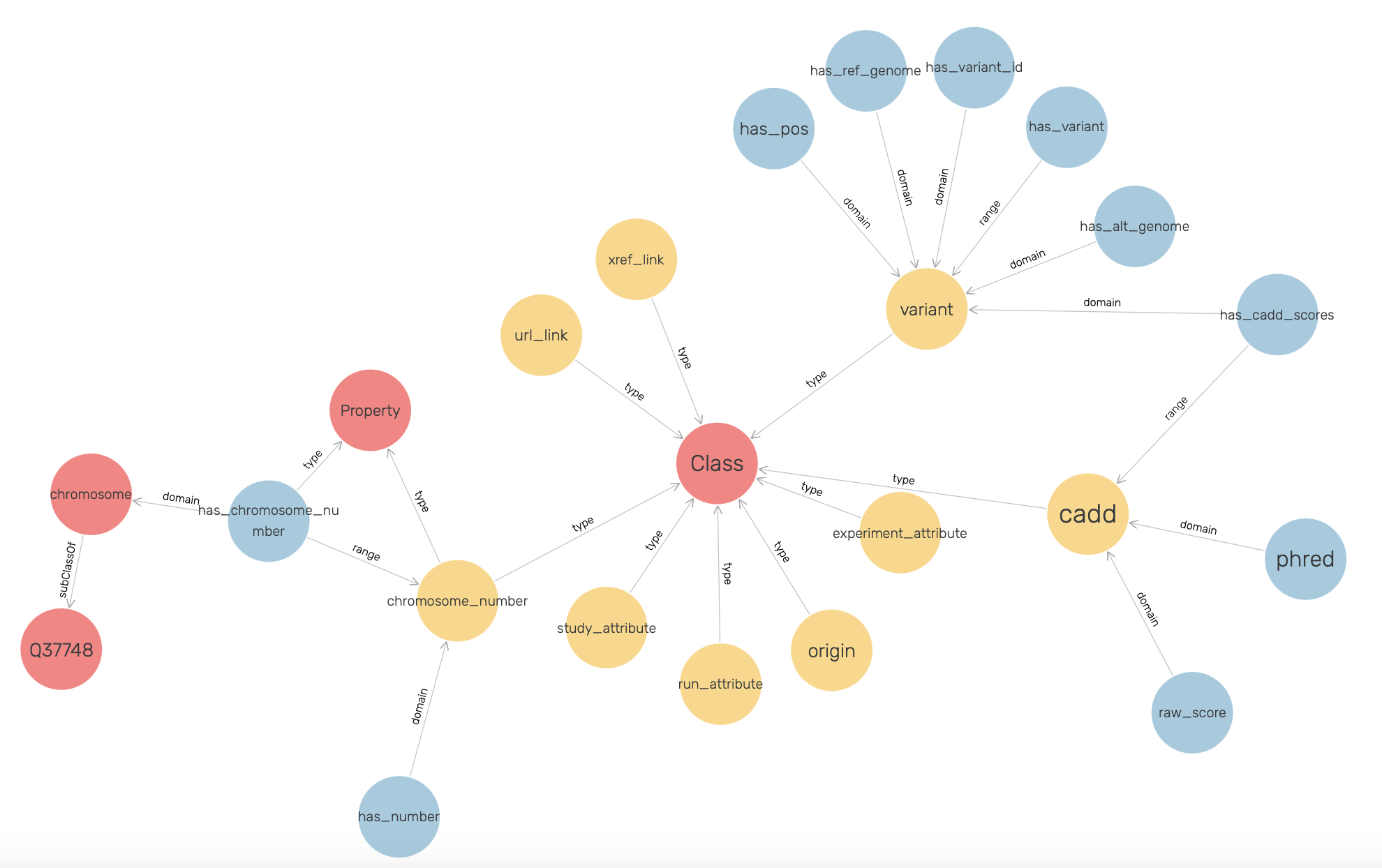}
    \caption{Ontology of the knowledge graph.}
    \label{fig:ontology}
\end{figure} 

A knowledge graph is represented using an ontology, which can be represented using a formal language such as RDF (Resource Description Framework), OWL (Web Ontology Language), or another domain-specific language. The ontology in this work has been represented using RDF. Each node-edge-node is represented as a triple by RDF. In a triple, the subject defines the first node, and the object defines the second node. The predicate defines the edge or relation joining the two nodes. A triple always ends with a dot. 

An ontology mainly consists of classes, properties, and relationships. In Figure \ref{fig:ontology}, the sub-classes are depicted by red nodes, the classes are by yellow nodes, and the relations by blue nodes. The description of the classes has been given in Table \ref{tab:classes-description} and the description of the properties has been given in Table \ref{tab:properties-description}.

\begin{table}[!h]
    \centering
    \caption{Description of the classes in the ontology.}
    \label{tab:classes-description}
    \begin{tabular}{|lp{30em}|}
        \hline
        CLASS & DEFINITION  \\
        \hline
        Chromosome Number & Identifier of the chromosome; values can be `1', `2', …, `22', `X', `Y', `MT' \\
        Origin & Unique identifier of the variant annotated by SPARQLing Genomics tool \\
        Variant & Encapsulates the different types of genomic alterations that can occur \\
        CADD & Encapsulates the different types of scores that can occur \\
        xref\_link & Type of annotation that provides a link between different resources or databases \\
        url\_link & Access link to experiment label \\
        study\_attribute & Metadata that describes the experimental design, data processing, and other aspects of a sequencing study \\
        run\_attribute & Metadata that describes the sequencing run \\
        experiment\_attribute & Metadata that describes the overall experimental design and goal of the experiment \\
        \hline
    \end{tabular}
\end{table}

\begin{table}[!h]
    \centering
    \caption{Description of the properties in the ontology.}
    \label{tab:properties-description}
    \begin{tabular}{|lp{15em}p{6.5em}p{5em}|}
        \hline
        PROPERTY & DEFINITION & DOMAIN & RANGE \\
        \hline
        has\_pos & Variant position & Variant & Integer \\
        has\_ref\_genome & Reference genome at that position & Variant & String \\
        has\_alt\_genome & Alternate genome at that position & Variant & String \\
        has\_variant\_id & Unique identifier of the variant & Variant & String \\
        has\_variant & Unique name given to the variant & Variant & String \\
        has\_cadd\_scores & Variant has associated CADD Scores & Variant & CADD \\
        has\_chromosome\_number & Chromosome has a chromosome number & Chromosome & String \\
        phred & Phred-scaled score & CADD & Long \\
        raw\_score & Raw CADD Score & CADD & Long \\
        \hline
    \end{tabular}
\end{table}

In the defined ontology, chromosome and variant are both domain classes, and a chromosome has an associated chromosome number to connect all similar chromosomes as an extension, and a variant has an associated variant ID. A variant has a reference and alternate genome. 

The ontology also explicitly defines CADD as a class where a variant has CADD Scores represented by both raw score and Phred-scale score, as properties of the CADD class. The ontology description has been given in Table \ref{tab:ontology-description}.

\begin{table}[!h]
    \centering
    \caption{Domain, properties, and ranges for the ontology.}
    \label{tab:ontology-description}
    \begin{tabular}{|lp{9em}p{10em}p{7em}|}
        \hline
        ENTITY & RDF:PROPERTY & DOMAIN & RANGE \\
        \hline
        Chromosome & Type & N/A & Wiki:Q37748 \\
         & SubClassOf & N/A & Wiki:Q37748 \\
        \hline
        has\_chromosome\_number & Type & N/A & Property \\
         & Domain & Chromosome & N/A \\
         & Range & chromosome\_number & N/A \\
        \hline
        chromosome\_number & Type & N/A & Class \\
        \hline
        has\_number & Type & N/A & Property \\
         & Domain & chromosome\_number & N/A \\ 
         & Range & xsd:int & N/A \\ 
        \hline
        Variant & Type & N/A & Class \\
        \hline
        has\_variant & Type & N/A & Property \\ 
         & Domain & Chromosome & Variant \\ 
         & Range & Variant & N/A \\
        \hline
        has\_pos & Type & N/A & Property \\ 
         & Domain & Variant & xsd:string \\ 
         & Range & xsd:int & N/A \\
        \hline
        has\_ref\_genome & Type & N/A & Property \\
         & Domain & Variant & xsd:string \\
         & Range & xsd:string & N/A \\
        \hline
        has\_alt\_genome & Type & N/A & Property \\
         & Domain & Variant & xsd:string \\
         & Range & xsd:string & N/A \\
        \hline
        CADD & Type & N/A & Class \\
        \hline
        has\_cadd\_score & Type & N/A & Property \\
         & Domain & Variant & CADD \\
         & Range & CADD & N/A \\
        \hline
        raw\_score & Type & N/A & Property \\
         & Domain & CADD & xsd:long \\
         & Range & xsd:long & N/A \\
        \hline
        phred & Type & N/A & Property \\
         & Domain & CADD & xsd:long \\
         & Range & xsd:long & N/A \\
        \hline
    \end{tabular}
\end{table}

\subsection{Conversion of VCF files to Knowledge Graphs}
\label{sec3.5:vcf-to-kg}

To transform the data in VCF, SPARQLing Genomics \citep{di2018sparqling} was utilized. SPARQLing Genomics is an open-source platform for querying and analyzing genomic data using the Semantic Web and Linked Data technologies. The platform provides an easy-to-use interface as well, that has been built to support SPARQL queries and various SPARQL query features, including sub-queries, filters, and aggregates. SPARQLing Genomics provides several in-built, ready-to-use tools, one of which is vcf2rdf that converts VCF data into RDF triples. 

The triples generated by the tool consist of uniquely identifiable names with symbolic and literal values like numbers or text. 

\begin{lstlisting}
    #CHROM POS   ID REF ALT QUAL    FILTER    INFO sample
       1  16963  .   G   A  45.64 SnpCluster AC=1;AF=0.500;AN=2;BaseQRankSum=1.465;DP-8;ExcessHet=3.0103;FS=0.000;MLEAC=1;MLEAF=0.500;MQ=60.00;MQRankSum=0.000;QD=5.70;ReadPosRankSum=-0.366;SOR=0.169GT:AD:D:GQ:PL0/1:6,2:8:53:53,0,228
\end{lstlisting}

The following is an example of how a variant position (example shown above) is translated into a triple by the tool.

\begin{lstlisting}
    <origin://4a37140cdc877d90ffe258a8151f27e@0> <http://biohackathon.org/resource/faldo#position> "16963"^^<http://www.w3.org/2001/XMLSchema#integer> .
\end{lstlisting}

As seen in the above example, the variant position is described with Feature Annotation Location Description Ontology (FALDO) \citep{bolleman2016faldo}. For other features not defined by FALDO, the URI is customized to the tool.

Each VCF file eventually corresponds to one large knowledge graph that was originally stored in an N3 format. N3 format is one of several formats supported by RDF and can be considered a shorthand non-XML serialization of RDF models. However, to accommodate the accessionID that would map to an unidentified patient, the N3 serialization was converted to NQ format with the accessionID as the named graph. An example of a triple from an N3 file has been given here.

\begin{lstlisting}
    <origin://4a37140cdc877d90ffe2b58a8151f27e@0> <sg://0.99.11/vcf2rdf/variant/REF> <sg://0.99.11/vcf2rdf/sequence/G>  .
\end{lstlisting}

The triple was then converted to NQ format, which yielded the following triple:

\begin{lstlisting}
    <origin://4a37140cdc877d90ffe2b58a8151f27e@0>   <sg://0.99.11/vcf2rdf/variant/REF> <sg://0.99.11/vcf2rdf/sequence/G> <sg://SRR13112995>.
\end{lstlisting}

The ontology was extended to accommodate the new relations generated by the tool. 

\subsection{Conversion of CADD Score files to Knowledge Graphs}
\label{cadd-to-kg}

The SnpEff and vcf2rdf tools were useful for converting VCF files to triples. However, the CADD Scores obtained through the pipeline were in tab-separated (TSV) format. To enrich the knowledge graphs, the CADD Scores had to be translated to RDF triples as well. Therefore, the ontology for CADD Scores was explicitly defined. 

\begin{figure}[H]
    \centering
    \includegraphics[keepaspectratio, width=0.7\textwidth, height=0.3\textheight]{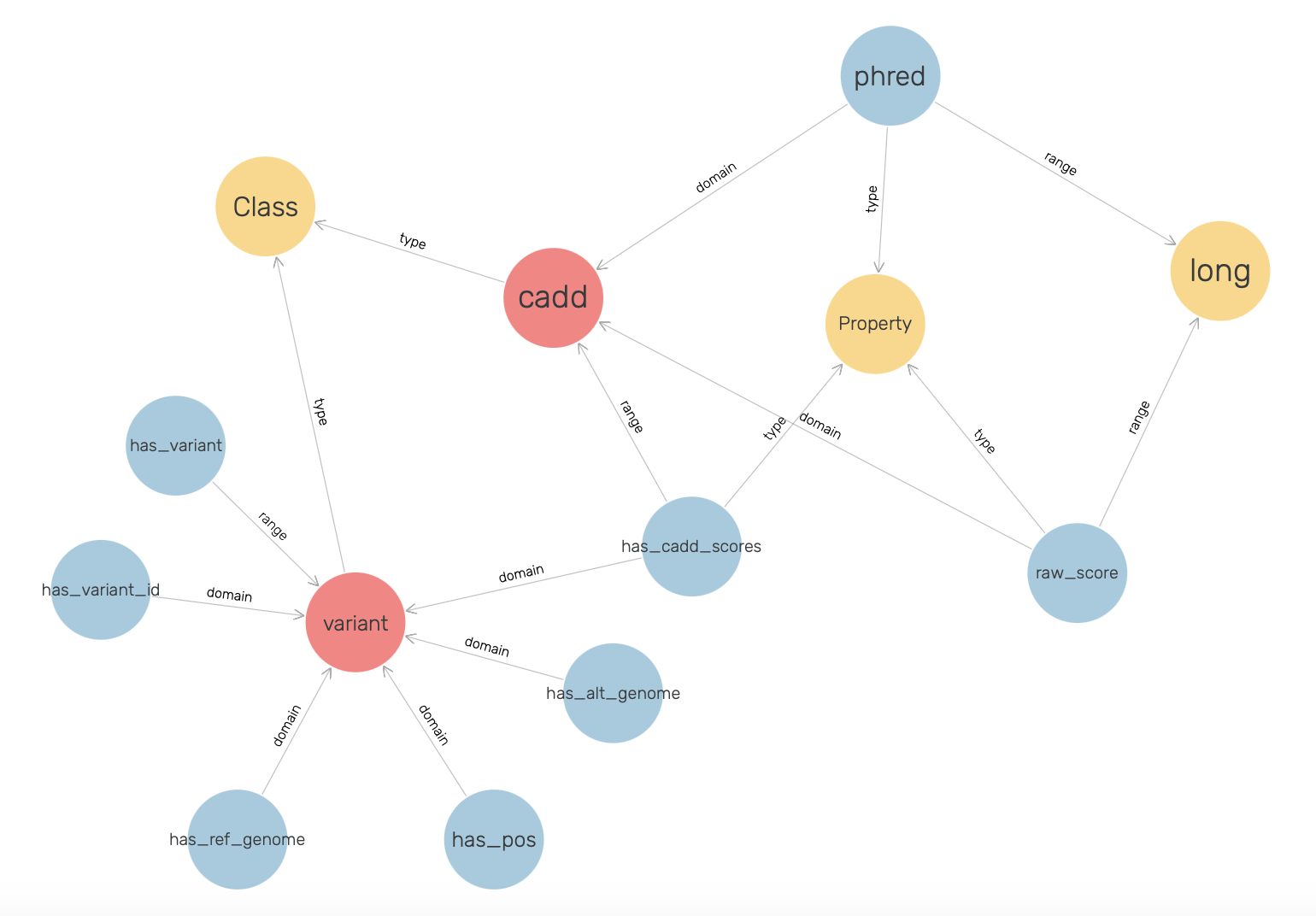}
    \caption{Ontology for CADD Scores.}
    \label{fig:cadd-scores-ontology}
\end{figure}

To visualize the graph, GraphDB has been utilized, and the ontology that was defined for CADD Scores has been shown in Figure \ref{fig:cadd-scores-ontology}

These scores have been represented with respect to the fields described in the VCF files, such as the chromosome, position, reference genome, and alternate genome. The raw scores and Phred-scale scores were obtained from the original TSV files.

The following is an example of a record in a TSV file for which the raw and phred scores map to chromosome 1 with position 16963, reference genome `G' and alternate genome `A' in the VCF file:

\begin{table}[H]
    \centering
    \label{tab:ex-cadd}
    \begin{tabular}{|lp{5em}p{5em}p{5em}p{5em}p{5em}|}
        \hline
        \#Chrom & Pos & Ref & Alt & RawScore & PHRED \\
        \hline
        1 & 16963 & G & A & 0.900784 & 12.72 \\
        \hline
    \end{tabular}
\end{table}

Each data record, like the above example, was converted to a Turtle triple (TTL), another format supported by RDF. A TTL format writes a graph in a compact textual form. There are only 3 parts to this triple – subject, predicate, and object. An example of the above data record converted to a TTL triple is given below:

\begin{lstlisting}
   <http://sg.org/SRR13112995/1/variant1> a ns1:variant ;
        ns1:has_alt_genome "A" ;
        ns1:has_cadd_scores <http://sg.org/SRR13112995/1/variant1/cadd> ;
        ns1:has_pos 16963 ;
        ns1:has_ref_genome "G" .
\end{lstlisting}


Each VCF file eventually corresponds to one large knowledge graph that was originally stored in an N3 format. N3 format is one of several formats supported by RDF and can be considered a shorthand non-XML serialization of RDF models. However, to accommodate the accessionID that would map to an unidentified patient, the N3 serialization was converted to NQ format with the accessionID as the named graph. An example of a triple from an N3 file has been given here.

\begin{lstlisting}
    <origin://4a37140cdc877d90ffe2b58a8151f27e@0> <sg://0.99.11/vcf2rdf/variant/REF> <sg://0.99.11/vcf2rdf/sequence/G>  .
\end{lstlisting}

The triple was then converted to NQ format, which resulted in the following triple:

\begin{lstlisting}
    <origin://4a37140cdc877d90ffe2b58a8151f27e@0>   <sg://0.99.11/vcf2rdf/variant/REF> <sg://0.99.11/vcf2rdf/sequence/G> <sg://SRR13112995>.
\end{lstlisting}

The ontology was extended to accommodate the new relations generated by the tool. 

\section{Graph Storage And Database}
\label{graph-storage-database}

Each variant file represented a single knowledge graph, so to unify several knowledge graphs into one single large graph, BlazeGraph~\cite{blazegraph} has been leveraged. This large knowledge graph was then queried to create a dataset for a case study. It is important to note that the edges are homogeneous in nature.

BlazeGraph is a high-performance, horizontally scalable, and open-source graph database that can be used to store and manage large-scale graph data. It has been designed to provide efficient graph querying and supports RDF data model that allows it to store and process both structured and semi-structured data. BlazeGraph uses a distributed architecture that can be easily integrated with other big data tools, such as Hadoop and Spark, to perform complex analytics on large-scale graph data.

BlazeGraph has been leveraged for efficiently querying the knowledge graphs to generate the dataset for Graph Neural Network downstream tasks. Other tools such as RIQ \cite{RIQ2017, RIQ2016, RIQ2015} can be used to index and query RDF-named graphs.

The total number of triples in the knowledge graph, after aggregating only 511 VCF files on a single machine, is as large as 3.1 Billion.

\section{KG Inference: Case Study}
\label{kg-inference}

In this section, we will discuss the usage of the knowledge graph created and demonstrate the use of VariantKG. We aim to use the KG for a classification task using graph machine learning. For the classification task, we are leveraging the open-source graph-based library called Deep Graph Library \cite{wang2019dgl}.

We packaged the several components discussed above in a user-friendly, robust and efficient tool called VariantKG, to empower users to train their own GNN models using our data or a combination of our data and new VCF files. We will further expand on the workings of our tool in Section \ref{kg-inference-variant-kg}.

\subsection{Deep Graph Library}
\label{kg-inference-dgl}

To implement the task, Deep Graph Library (DGL), an open-source library supporting graph-based deep learning was utilized. DGL provides a set of high-level APIs for building scalable and efficient graph neural network models. With DGL, we can create, manipulate, and learn from large-scale graphs with billions of nodes and edges.

There are three main tasks supported by DGL:
\begin{itemize}
    \item Node Classification: Predict the class or label of a node in a graph based on its features.
    \item Link Prediction: Predict if there is a connection or an edge between two nodes.
    \item Graph Classification: Classify an entire graph into one or more classes or categories.
\end{itemize}

DGL represents a graph as a DGLGraph object, which is a framework-specific graph object. It requires the number of nodes and a list of source and destination nodes, where nodes and edges must have consecutive IDs starting from 0. Since DGL only accepts numeric input, all strings such as URI were mapped to integers. In this case study, node classification was used to classify variants into CADD Score categories based on their features.

\subsection{Node Classification Task}
\label{kg-inference-node-classification}

For this task, Graph Convolutional Network (GCN) \cite{zhang2019graph} and GraphSAGE \cite{hamilton2017sage} have been used. For both models, each node is associated with a feature vector.

GCNs use node embeddings and adjacency matrices to compute new embeddings while training. Similar to CNN, the model weights and biases are first initialized to 1, and then a section of the graph is passed through the model. A non-linear activation function is used to compute predicted node embeddings for each node. Cross entropy loss is calculated to quantify the difference between the predicted node embeddings and the ground truth. Loss gradients are then computed to update the model using the ADAM optimizer\cite{kingma2014adam} for this task. These steps are repeated until convergence.

GraphSAGE uses SAGEConv layers where for every iteration, the output of the model involves finding new node representation for every node in the graph. Mean is used as the aggregation function, and the ReLU activation function has been utilized. ADAM optimizer was used for this model as well. One of the most noted properties of GraphSAGE is its ability to aggregate neighbor node embeddings for a given target node. Through the experiments conducted, this property was observed. GraphSAGE also generalizes better to unseen nodes because of its ability to perform inductive learning on graphs. 

\begin{figure}[H]
    \centering
    \includegraphics[width=0.6\textwidth, height=0.2\textheight]{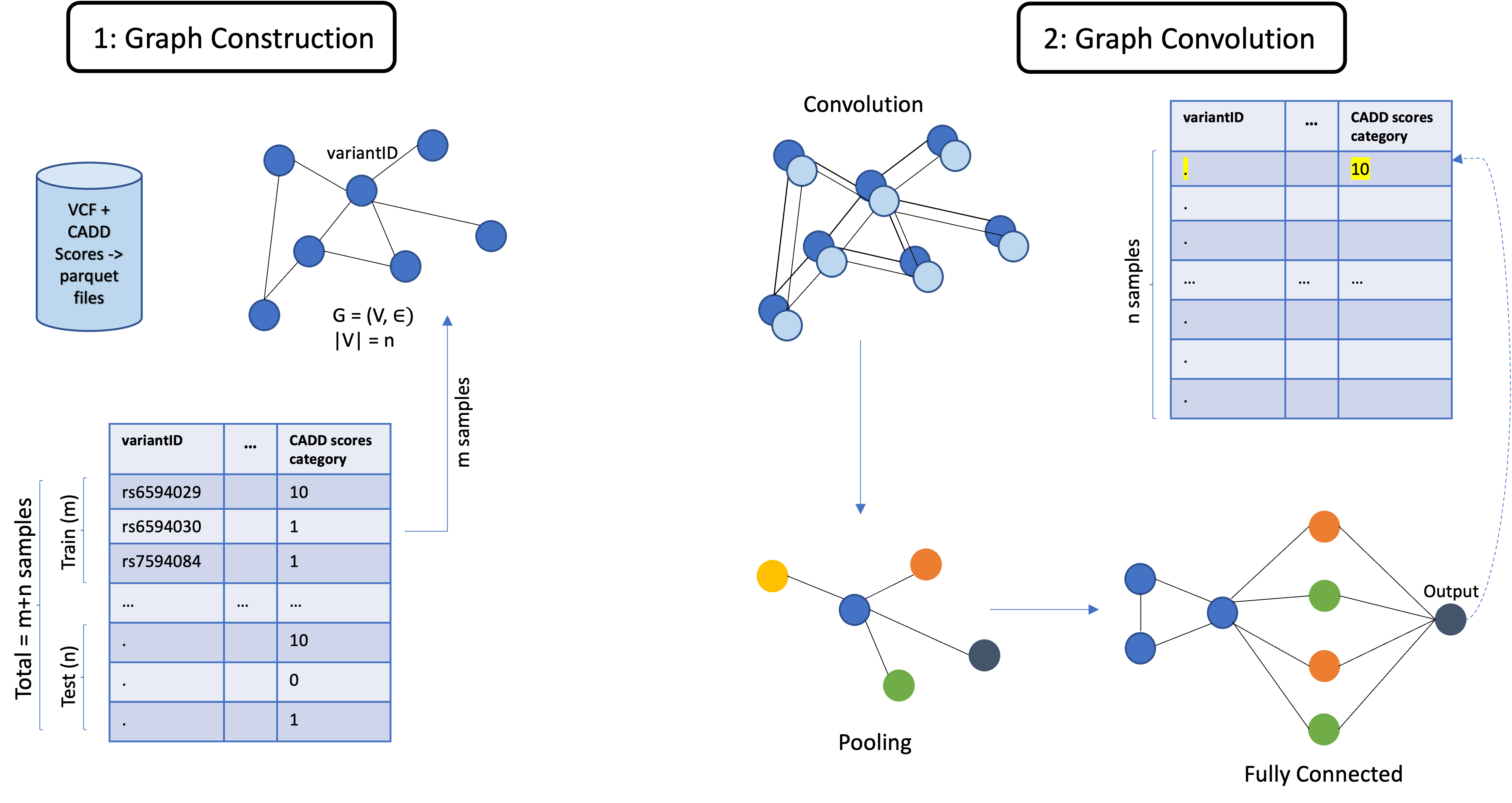}
    \caption{Architecure of GCN.}
    \label{fig:graph-conv-net}
\end{figure}

\begin{figure}[H]
    \centering
    \includegraphics[width=0.6\textwidth, height=0.2\textheight]{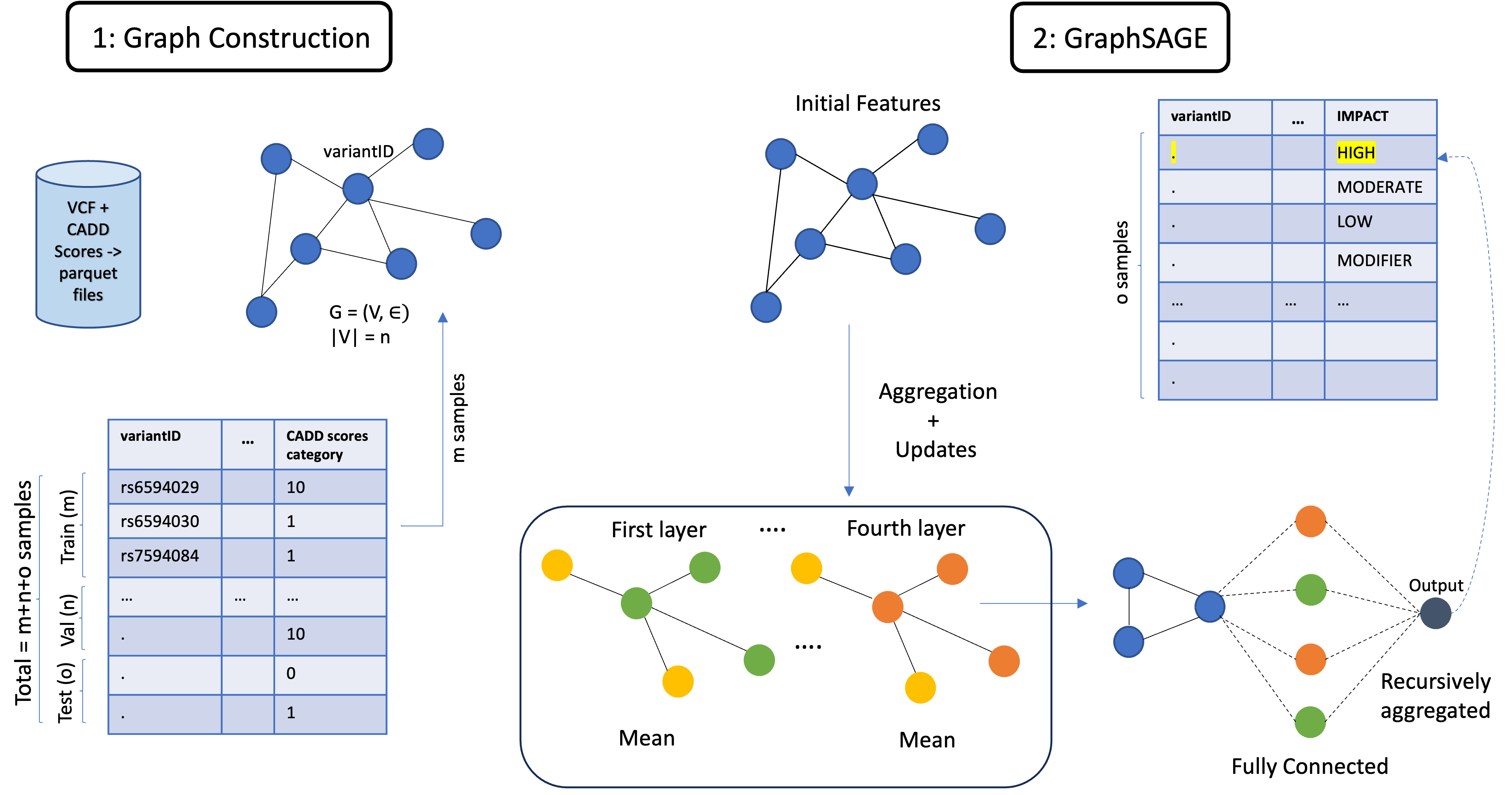}
    \caption{Architecure of GraphSAGE.}
    \label{fig:graphsage}
\end{figure}

The architecture of GCN has been shown in Figure \ref{fig:graph-conv-net}. The architecture of GraphSAGE, shown in Figure \ref{fig:graphsage}, differs in the property of message passing between the nodes. This was crucial as the nodes in the input graph relied on several pieces of information from their neighboring nodes.

\subsection{Experiments \& Results}
\label{kg-inference-experiments-results}

\subsubsection{Experiment Setup}
\label{kg-inference-experiment-setup}

The experiments were run on CloudLab \cite{ricci2014introducing}, a testbed for cloud computing research and new applications. CloudLab cluster at Clemson University was built in partnership with Dell. Clemson machines were used to carry out the standalone training experiments. They were carried out on nodes with 16 cores per node (2 CPUs), 12x4 TB disk drives in each node, plus 8x1 TB disks in each node. The nodes are configured with 256 GB of memory and 73TB of RAM. The operating system installed across all of them was Ubuntu 18.04.

Wisconsin machines were also used to carry out the standalone training experiments described in section 6.2. The CloudLab cluster at the University of Wisconsin was built in partnership with Cisco, Seagate, and HP. The cluster has 523 servers with a total of 10,060 cores and 1,396 TB of storage, including SSDs on every node. 

The nodes were chosen upon availability for the standalone node classification tasks.

\subsection{VariantKG as a tool}
\label{kg-inference-variant-kg}

\begin{figure}[H]
  \centering
  \includegraphics[keepaspectratio, width=\textwidth, height=0.5\textheight]{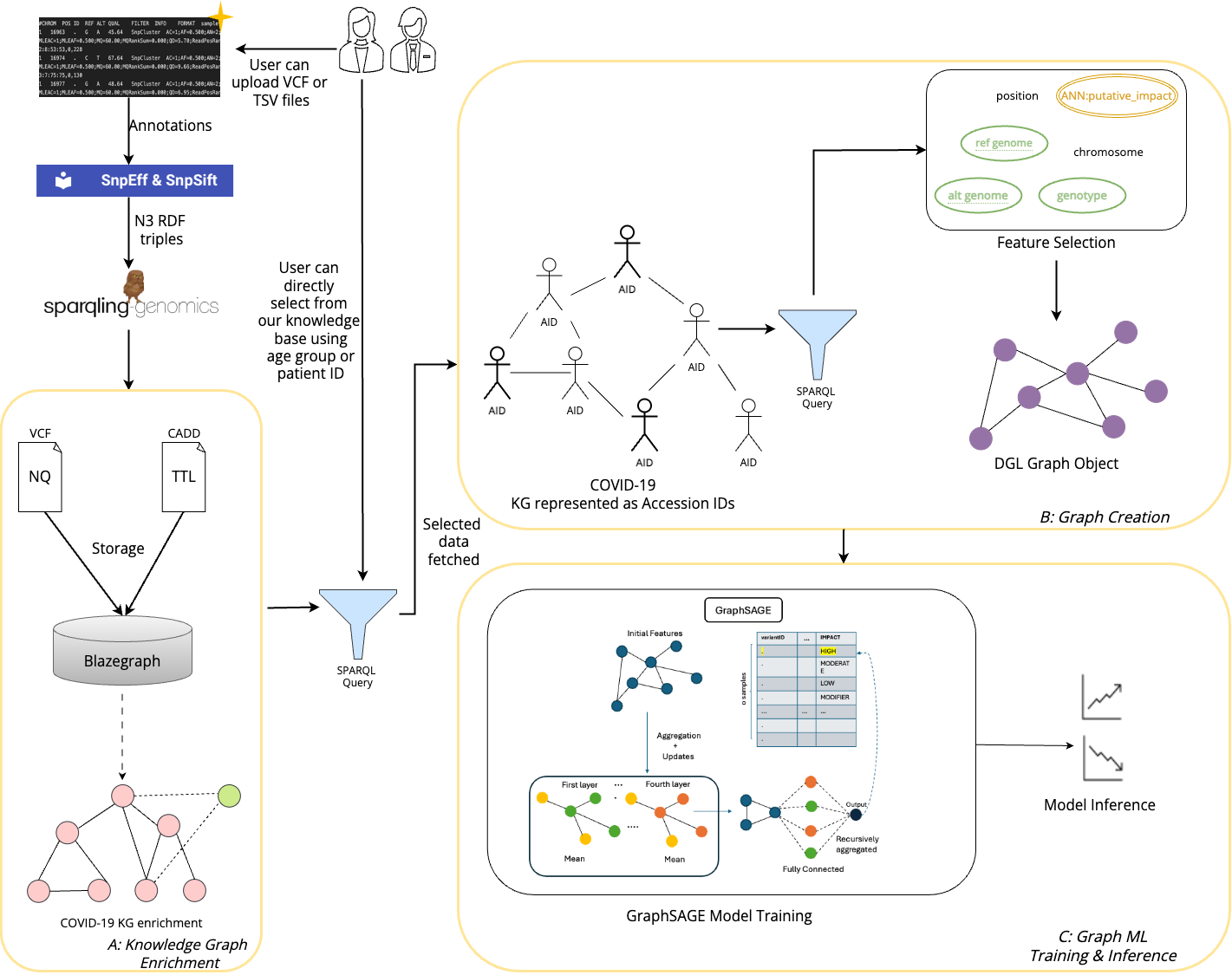}
  \caption{Architecture of VariantKG.}
  \label{fig:arch}
\end{figure}

The architecture of our tool, shown in Figure \ref{fig:arch} has been designed to facilitate the extraction, processing, and analysis of variant-level genomic data using GNNs. The workflow integrates custom data inputs, feature selection, storage in a graph database, graph creation, and model training for inferencing genomic information. Given the rapidly expanding repository of genomic data, VariantKG offers a robust platform for researchers or users to extract relevant information and train Graph Neural Networks (GNNs) such as GraphSAGE for inferential purposes. Users have the flexibility to upload one or more VCF files, which may include associated CADD Scores for specific variants. Alternatively, users can select from pre-existing datasets that have been structured into a large-scale, comprehensive knowledge graph.

The first part of our tool, shown as \textit{A: Graph Enrichment} handles data upload and pre-processing, where users can upload one or more VCF with the associated CADD Scores or select from the pre-existing database that contains genomic variants and annotations. If the user chooses to upload the files, the VCF and CADD Scores files are processed using SnpEff to provide additional variant-level annotation and SPARQLing-Genomics for transforming the data into RDF format using the \textit{vcf2rdf} tool. The processed data is then stored along with the existing data on Blazegraph, which is a scalable graph database. The data from the VCF files is stored as NQ RDF-triples, and the data from the CADD Scores files is stored as TTL (turtle) RDF-triples. This data is essentially incorporated into our existing knowledge base, enriching the KG with new information. The ontology defined for the knowledge graphs allows for this extensibility. Users also have the option to select the accession IDs corresponding to de-identified patients from our existing large-scale knowledge graph. This feature has been provided by utilized an efficient, well-formed SPARQL query that fetches all the unique accession IDs that are available in the knowledge graph. This will be further discussed in \ref{scenario1:graph-enrichment}

The second part in the workflow of our tool, shown as \textit{B: Graph Creation} is the availability of feature selection and creation of the subgraph or graph suitable for model training. The user can select the features that are available in the NQ VCF files, which are passed through a well-formed SPARQL query that retrieves the selected features. The query results are temporarily stored in a columnar storage format optimized for analytical queries. The user can then assign a feature as the class label, facilitating supervised learning tasks such as node classification. Once the user selects one or more accession IDs from the knowledge graph, the user will then have an option to select the features based on the patient's variant-level information. The data along with the selected features and class labels is used to create a Deep Graph Library (DGL)-specific \textit{DGLGraph} object that represents the graph as integers. This is essential as DGL necessitates that the input data be formatted into a frame-work specific graph object. This will be further discussed in \ref{scenario2:graph-creation}

Once the graph has been constructed, users can then train the DGL GNN model, GraphSAGE and gain a better understanding of the model performance through inferencing, shown as \textit{C: Graph ML Training \& Inference}. As of now, we support only GraphSAGE but intend to extend the support to other GNN models in the future. The training process can be customized based on user-defined model hyperparameters. Existing data that have been prestructured as DGL graphs can also be loaded directly for model training. The user can observe the training process through the displayed charts and gain inferencing knowledge from the confusion matrices to assess the model performance. This will be further discussed in \ref{scenario3:graph-ml}

The UI was developed using Gradio \cite{abid2019gradio}, HTML5 and JavaScript, and the backend code was developed using Python 3.8 on Ubuntu 18.04.

\subsubsection{Scenario 1: Graph Enrichment}
\label{scenario1:graph-enrichment}

\begin{figure}[!h]
    \centering
    \includegraphics[keepaspectratio, width=\textwidth, height=0.5\textheight]{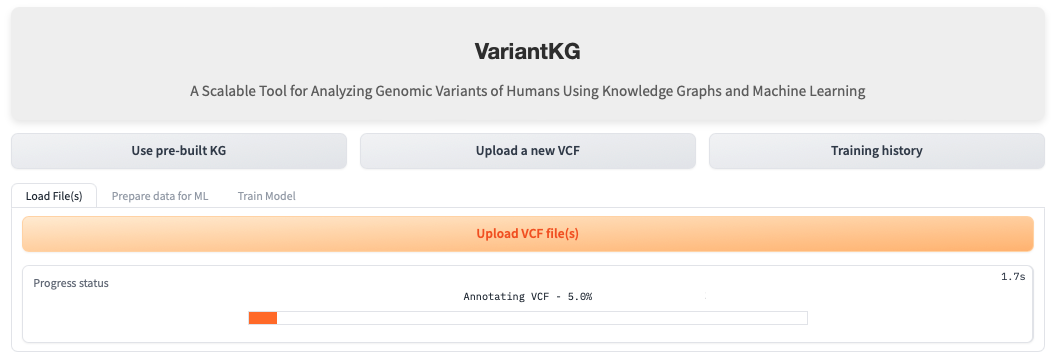}
    \caption{Graph Enrichment where user can upload VCF files to enrich the KG in the VariantKG tool.}
    \label{fig:variant_kg_graph_enrichment_upload}
\end{figure}

The first part of the VariantKG tool focuses on KG enrichment. A user can upload one or more VCF files containing variant-level information and CADD Scores in a TSV format as shown in Figures \ref{fig:variant_kg_graph_enrichment_upload} or the user can select patients from the existing KG. If the user uploads VCF files, the information in the files is first run through the SnpEFF tool. SnpEFF adds new annotated information to each variant in the VCF file and also updates the headers of the file to reflect the annotations. These annotations are functional information that is added to the `INFO' field using the `ANN' tag. This ANN field, in turn, consists of several bits of information, such as allele, annotation using Sequence and Ontology terms, putative impact, gene name, gene ID, feature type, feature ID, transcript biotype, exon or intron rank, cDNA position, protein position and several types of distances to the feature. 

Once the files have been annotated, we then utilize the \textit{vcf2rdf} tool provided by SPARQLing-Genomics to efficiently translate the information on each variant into several triples in an RDF-suitable N3 format. These N3 triples were converted to NQ format for several reasons. Converting to NQ format allows for the use of a named graph, which provides a robust way to group triples into distinct sets, considerably enhancing data organization and management. Another advantage of using named graph is the support it provides to easily attach new metadata to specific data subsets. Additionally, named graphs also enable more precise and efficient SPARQL queries, improving data extraction quality and speed. 

\begin{figure}[!h]
    \centering
    \includegraphics[keepaspectratio, width=\textwidth, height=0.5\textheight]{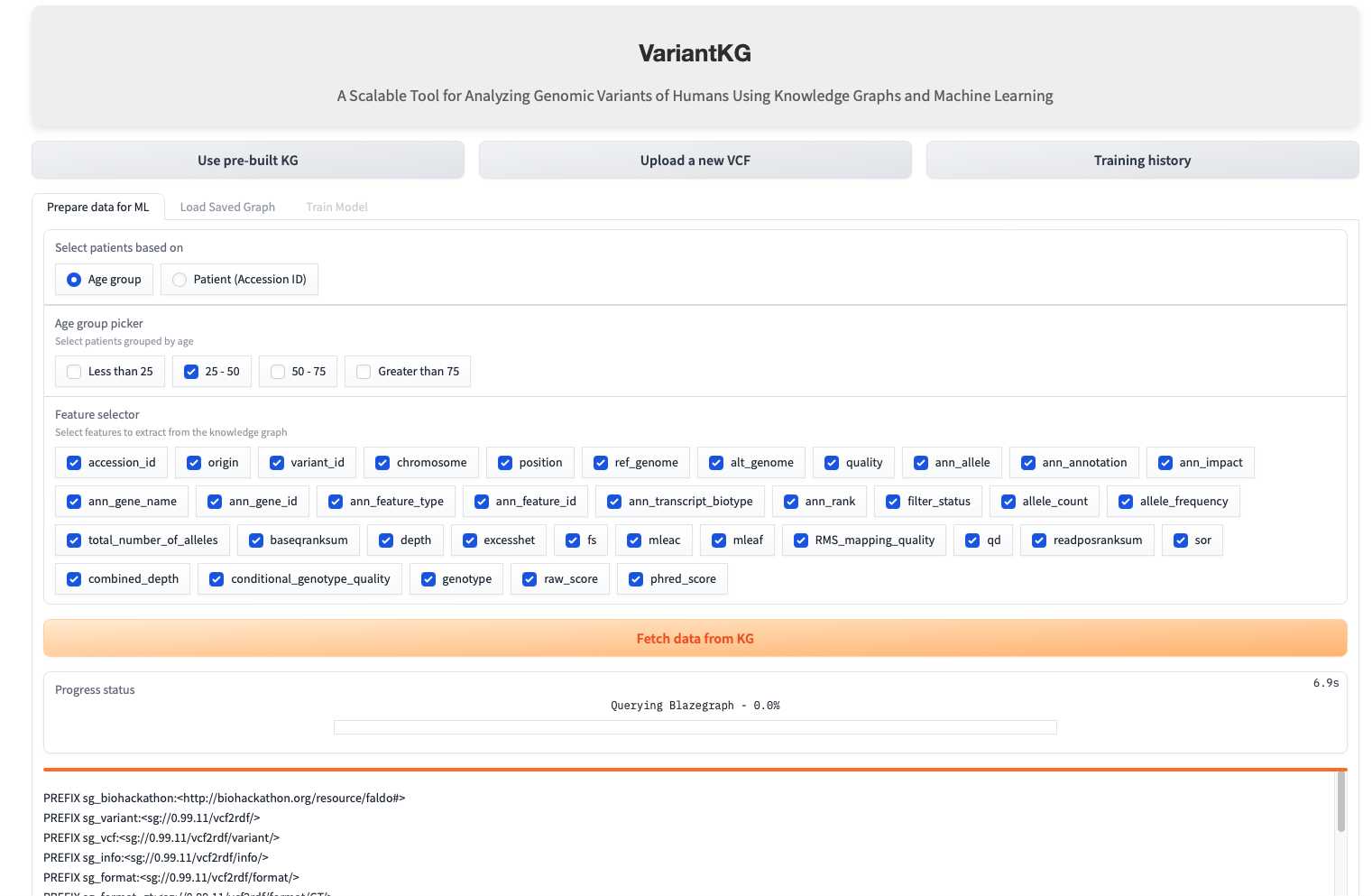}
    \caption{Graph Enrichment where user can select data from the KG based on age filter in VariantKG tool.}
    \label{fig:variant_kg_graph_enrichment_age_picker}
\end{figure}   

\begin{figure}[!h]
    \centering
    \includegraphics[keepaspectratio, width=\textwidth, height=0.5\textheight]{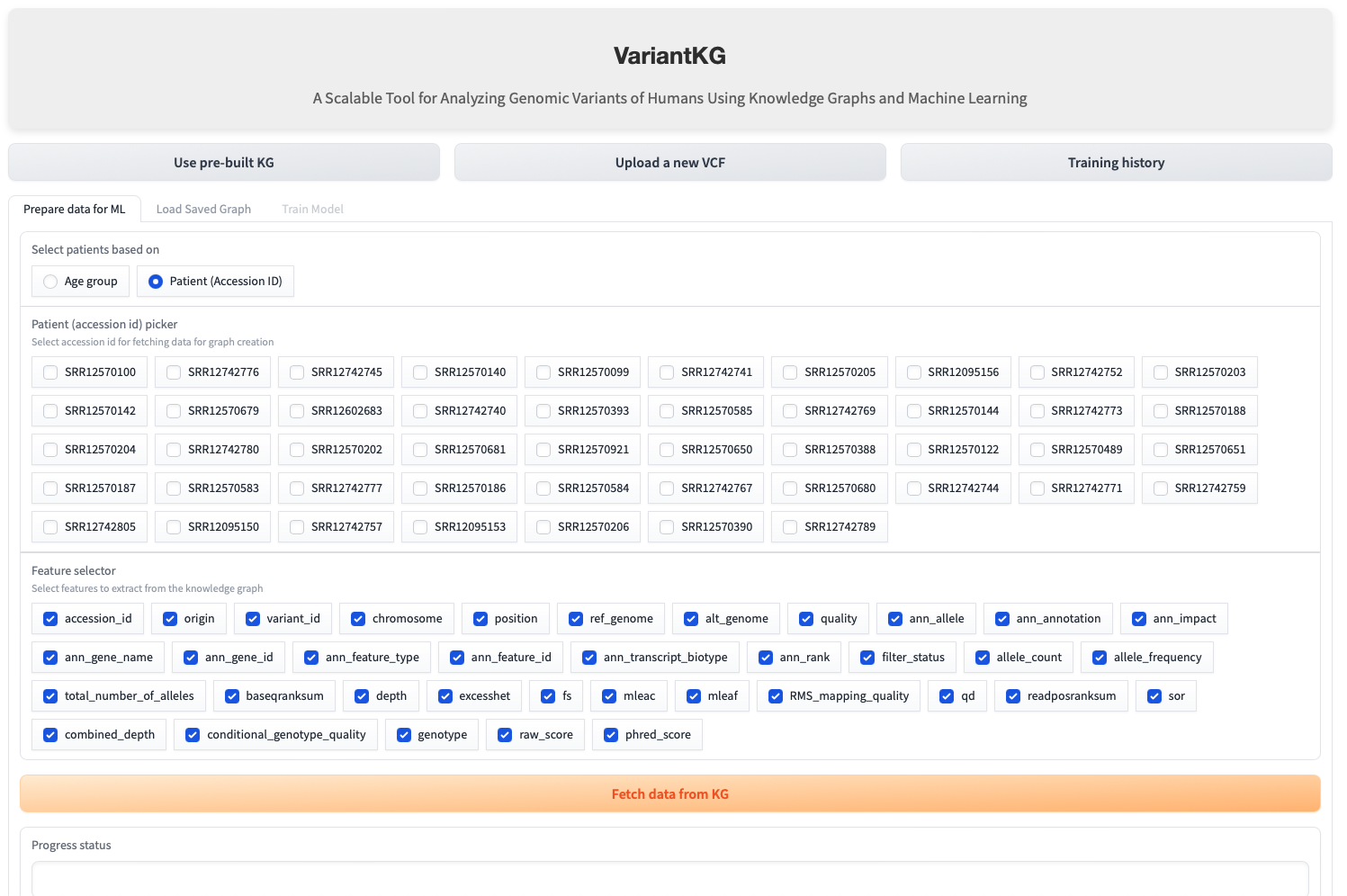}
    \caption{Graph Enrichment where user can select data from the KG based on existing patients in VariantKG tool.}
    \label{fig:variant_kg_graph_enrichment_accessionIDs}
\end{figure}

The triples are stored in Blazegraph, a high-performance graph database, which also stores the larger knowledge graph. To prepare the data from ML tasks, users can select patients using age groups as shown in Figure \ref{fig:variant_kg_graph_enrichment_age_picker}, or using the accession IDs, as shown in Figure \ref{fig:variant_kg_graph_enrichment_accessionIDs}, which are internally fetched using another SPARQL query that is only executed when the user wishes to prepare data for downstream ML tasks for efficiency. 

The user can then select the features from a list consisting of the original headers from the VCF files and the annotated features by SnpEff. Once the user hits the `Fetch from KG' button, another SPARQL query is then executed in the backend. If the user is selected from an age group, the SPARQL query consists of a filter that fetches all the accession IDs, which is passed as a list to the final query. If the user selects the accession IDs, it is passed as a list, similar to the previous query and the final query, shown below, fetches all the features selected for those patients or accession IDs.

\begin{lstlisting}
    PREFIX sg_biohackathon:<http://biohackathon.org/resource/faldo#>
    PREFIX sg_variant:<sg://0.99.11/vcf2rdf/>
    PREFIX sg_vcf:<sg://0.99.11/vcf2rdf/variant/>
    PREFIX sg_info:<sg://0.99.11/vcf2rdf/info/>
    PREFIX sg_format:<sg://0.99.11/vcf2rdf/format/>
    PREFIX sg_format_gt:<sg://0.99.11/vcf2rdf/format/GT/>
    PREFIX ns1:<http://sg.org/>
    
    SELECT DISTINCT ?accession_id ?origin (COALESCE(?variant_id, "None") AS ?variant_id) ?chromosome ?position ?ref_genome ?alt_genome ?quality ?ann ?ann_split_1 ?filter_status ?allele_count ?allele_frequency ?total_number_of_alleles ?baseqranksum ?depth ?excesshet ?fs ?mleac ?mleaf ?RMS_mapping_quality ?qd ?readposranksum ?sor ?combined_depth ?conditional_genotype_quality ?genotype ?raw_score ?phred_score
    WHERE {
      GRAPH ?accession_id {
        OPTIONAL { ?origin sg_variant:variantId ?variant_id . }
        BIND (COALESCE(?variant_id, "None") AS ?variant_id)
        ?origin sg_biohackathon:reference ?chromosome .
        ?origin sg_biohackathon:position ?position .
        ?origin sg_vcf:REF ?ref_genome .
        ?origin sg_vcf:ALT ?alt_genome .
        ?origin sg_vcf:QUAL ?quality .
        ?origin sg_info:ANN ?ann .
        BIND (IF(STRLEN(?ann) - STRLEN(REPLACE(?ann, ",", "")) = 0, ?ann, STRBEFORE(?ann, ",")) AS ?ann_split_1)
        OPTIONAL { ?origin sg_info:FILTER_STATUS ?filter_status . }
        OPTIONAL { ?origin sg_info:ALLELE_COUNT ?allele_count . }
        OPTIONAL { ?origin sg_info:ALLELE_FREQUENCY ?allele_frequency . }
        OPTIONAL { ?origin sg_info:TOTAL_NUMBER_OF_ALLELES ?total_number_of_alleles . }
        OPTIONAL { ?origin sg_info:BASEQRANKSUM ?baseqranksum . }
        OPTIONAL { ?origin sg_info:DEPTH ?depth . }
        OPTIONAL { ?origin sg_info:EXCESSHET ?excesshet . }
        OPTIONAL { ?origin sg_info:FS ?fs . }
        OPTIONAL { ?origin sg_info:MLEAC ?mleac . }
        OPTIONAL { ?origin sg_info:MLEAF ?mleaf . }
        OPTIONAL { ?origin sg_info:RMS_MAPPING_QUALITY ?RMS_mapping_quality . }
        OPTIONAL { ?origin sg_info:QD ?qd . }
        OPTIONAL { ?origin sg_info:READPOSRANKSUM ?readposranksum . }
        OPTIONAL { ?origin sg_info:SOR ?sor . }
        OPTIONAL { ?origin sg_info:COMBINED_DEPTH ?combined_depth . }
        OPTIONAL { ?origin sg_info:CONDITIONAL_GENOTYPE_QUALITY ?conditional_genotype_quality . }
        OPTIONAL { ?origin sg_info:GENOTYPE ?genotype . }
        OPTIONAL { ?origin sg_info:RAW_SCORE ?raw_score . }
        OPTIONAL { ?origin sg_info:PHRED_SCORE ?phred_score . }
      }
      ?origin sg_biohackathon:position ?position .
      OPTIONAL { ?variant <http://sg.org/has_pos> ?position . }
      ?origin sg_vcf:REF ?ref_genome .
      OPTIONAL { ?variant <http://sg.org/has_ref_genome> ?ref_genome . }
      ?origin sg_vcf:ALT ?alt_genome .
      OPTIONAL { ?variant <http://sg.org/has_alt_genome> ?alt_genome . }
      OPTIONAL { ?variant <http://sg.org/has_cadd_scores> ?cadd_scores . }
      OPTIONAL { ?cadd_scores <http://sg.org/has_raw_score> ?raw_score . }
      OPTIONAL { ?cadd_scores <http://sg.org/has_phred> ?phred_score . }
      FILTER (?accession_id IN (accession_id_list))
    } ORDER BY ?variant_id
\end{lstlisting}

\subsubsection{Scenario 2: Graph Creation}
\label{scenario2:graph-creation}

\begin{figure}[H]
    \centering
    \includegraphics[keepaspectratio, width=\textwidth, height=0.4\textheight]{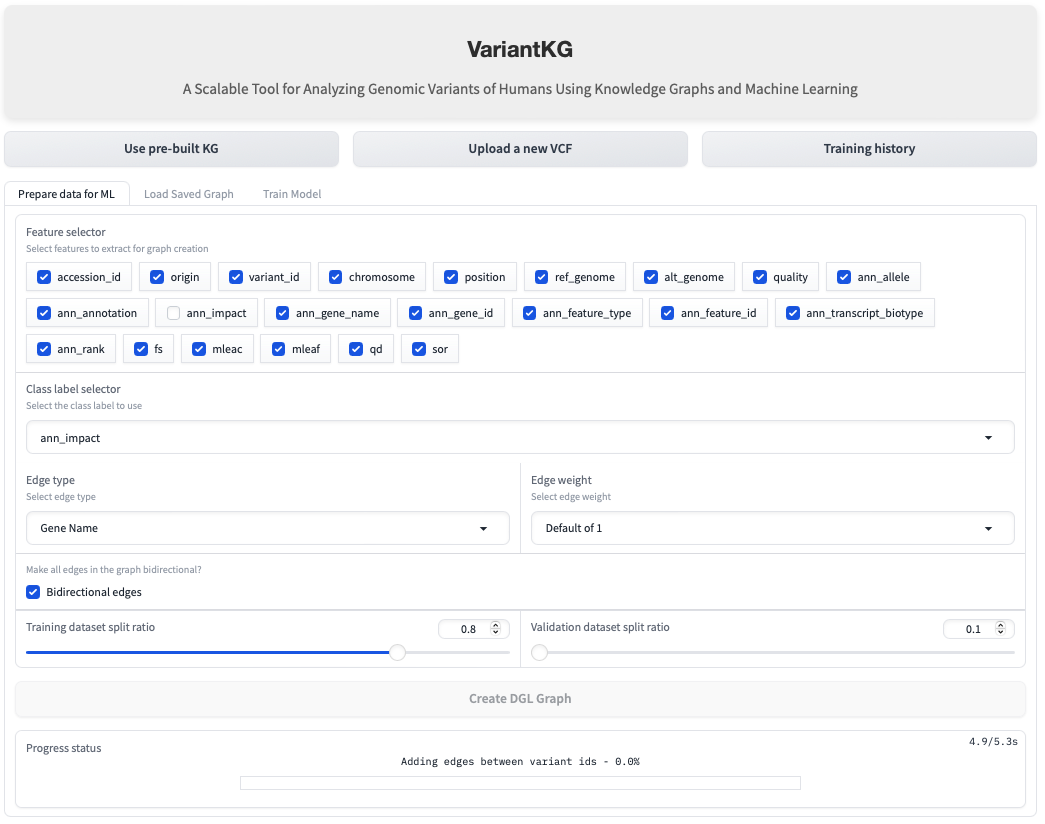}
    \caption{Graph Creation where user can prepare data for ML classification task in VariantKG tool.}
    \label{fig:variant_kg_graph_creation_prepare_data}
\end{figure}

Once the data has been fetched in the backend, the user is redirected to `Graph Creation' where the user can select node features in the graph and select the class label from a subset of the features that are meaningful to be classified as a part of the classification task. The user is then given complete control of the graph construction where the user can select the edge type which can be using Gene Name (default) or fully connected. The weight of the edges can be 1 (default), the number of incoming edges to a node, or a user-defined value. The user can then select if the edges should be bidirectional and the train:val splits. The default is 80:10 with the remaining 10\% calculated in the backend to reduce user clicks. Using this information, the graph is created for the node classification task. This is shown in Figure \ref{fig:variant_kg_graph_creation_prepare_data}

\subsubsection{Scenario 3: Graph Machine Learning \& Inference}
\label{scenario3:graph-ml}

\begin{figure}[!h]
    \centering
    \includegraphics[width=0.9\textwidth]{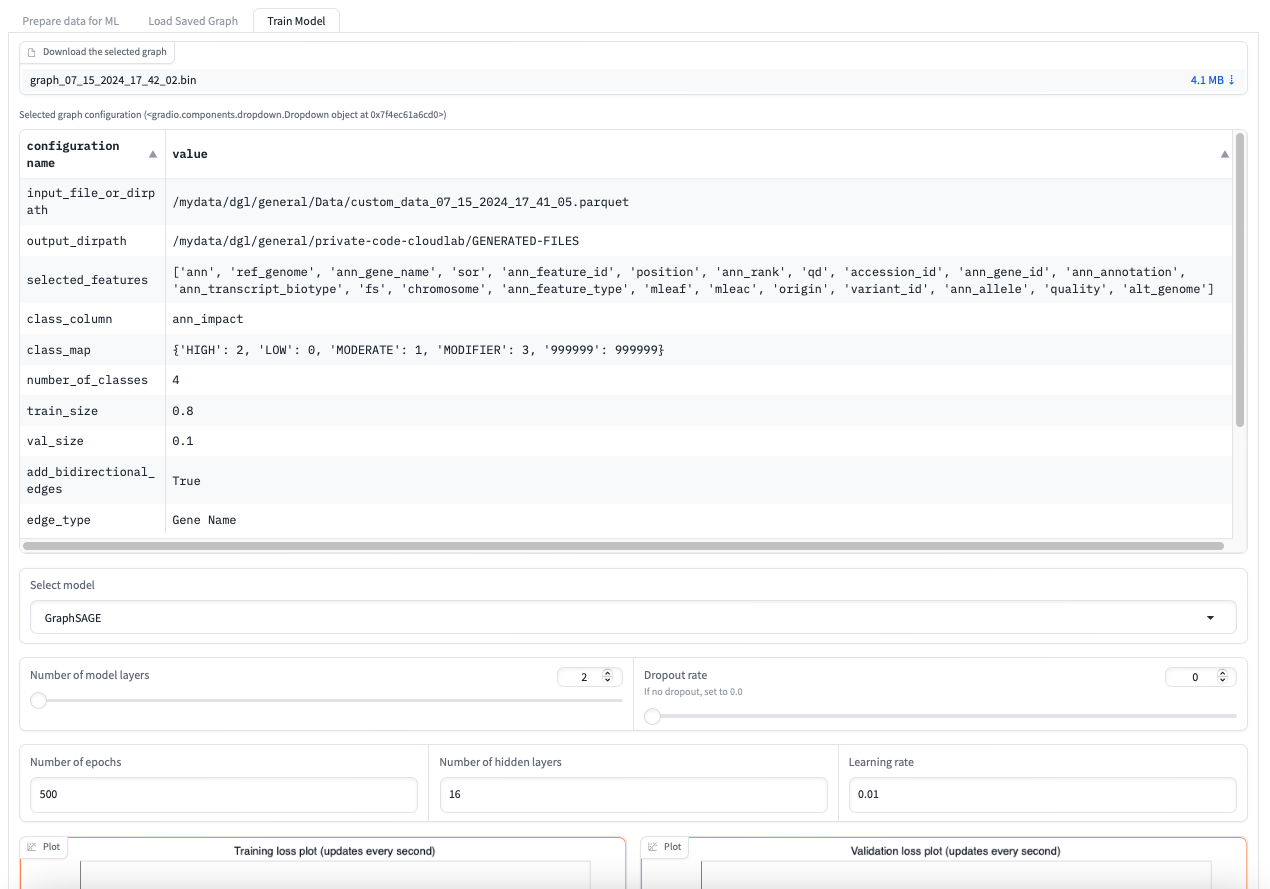}
    \caption{Graph ML Training in VariantKG tool.}
    \label{fig:variant_kg_graph_ML_parameter_setting}
\end{figure}

The graph for the node classification task needs to be in a DGL-specific input format. The data is thus converted to a `DGLGraph' object that consists of only integers that are mapped to the actual string values. The user can view the summary of the graph once created. This summary consists of all the selected features, the class label, the number of classes, graph properties, and the number of nodes and edges. The user is given a choice to download the DGL graph or continue with graph machine learning. For the ML task, the user can select between GraphSAGE and GCN models for the classification task and set the hyperparameters to train the model as shown in Figure \ref{fig:variant_kg_graph_ML_parameter_setting}. This includes the number of layers other than the input and output layers that constitute the primary architecture of the model, the number of hidden layers, the dropout rate, the learning rate, and the number of epochs for training. 

\begin{figure}[!h]
    \centering
    \includegraphics[width=0.9\textwidth]{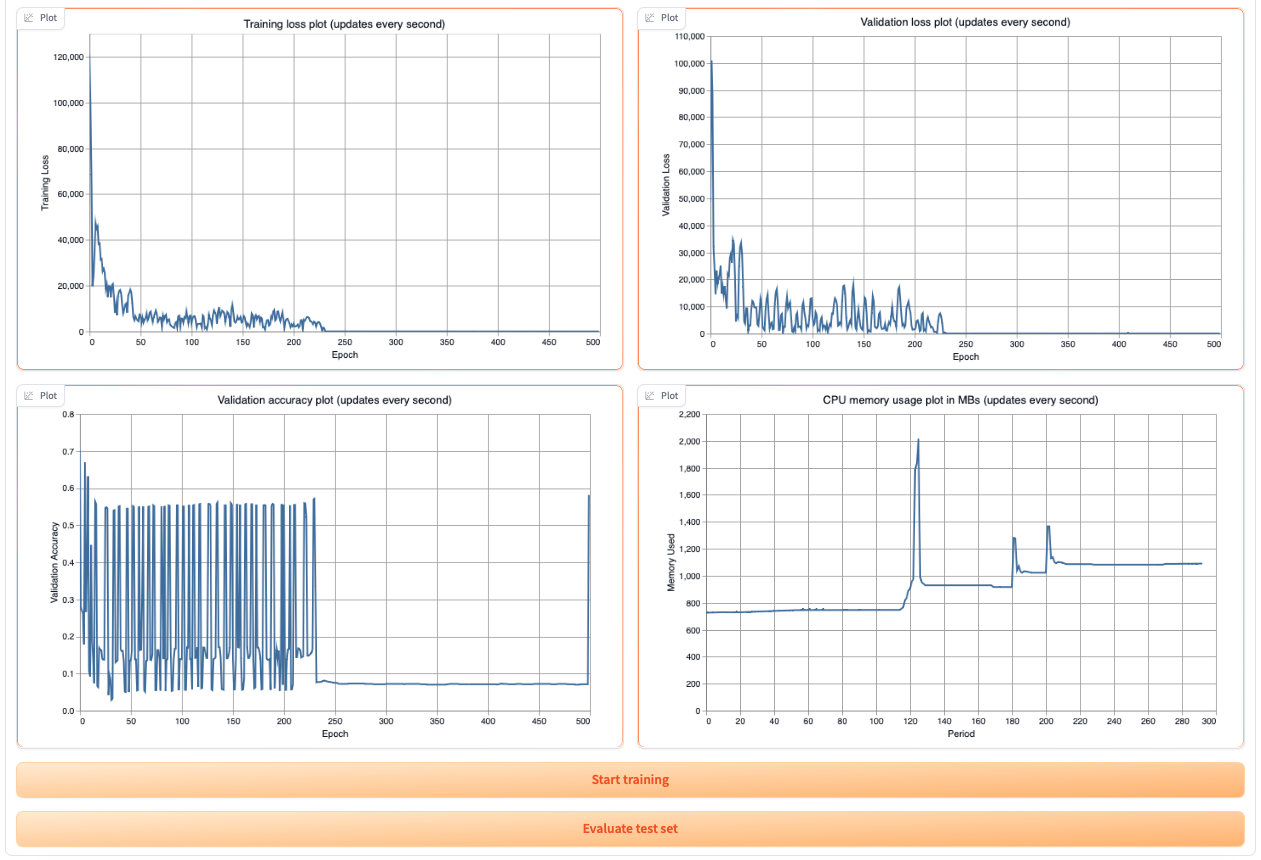}
    \caption{Graph ML Training and Validation loss, Validation accuracy and CPU usage plots in VariantKG tool.}
    \label{fig:variant_kg_graph_ML_training}
\end{figure}

Once the training begins, the user can view the training and validation loss plots, the validation accuracy plot, and the CPU memory usage as shown in Figure \ref{fig:variant_kg_graph_ML_training}. 

\begin{figure}[H]
    \centering
    \includegraphics[height=0.45\textheight]{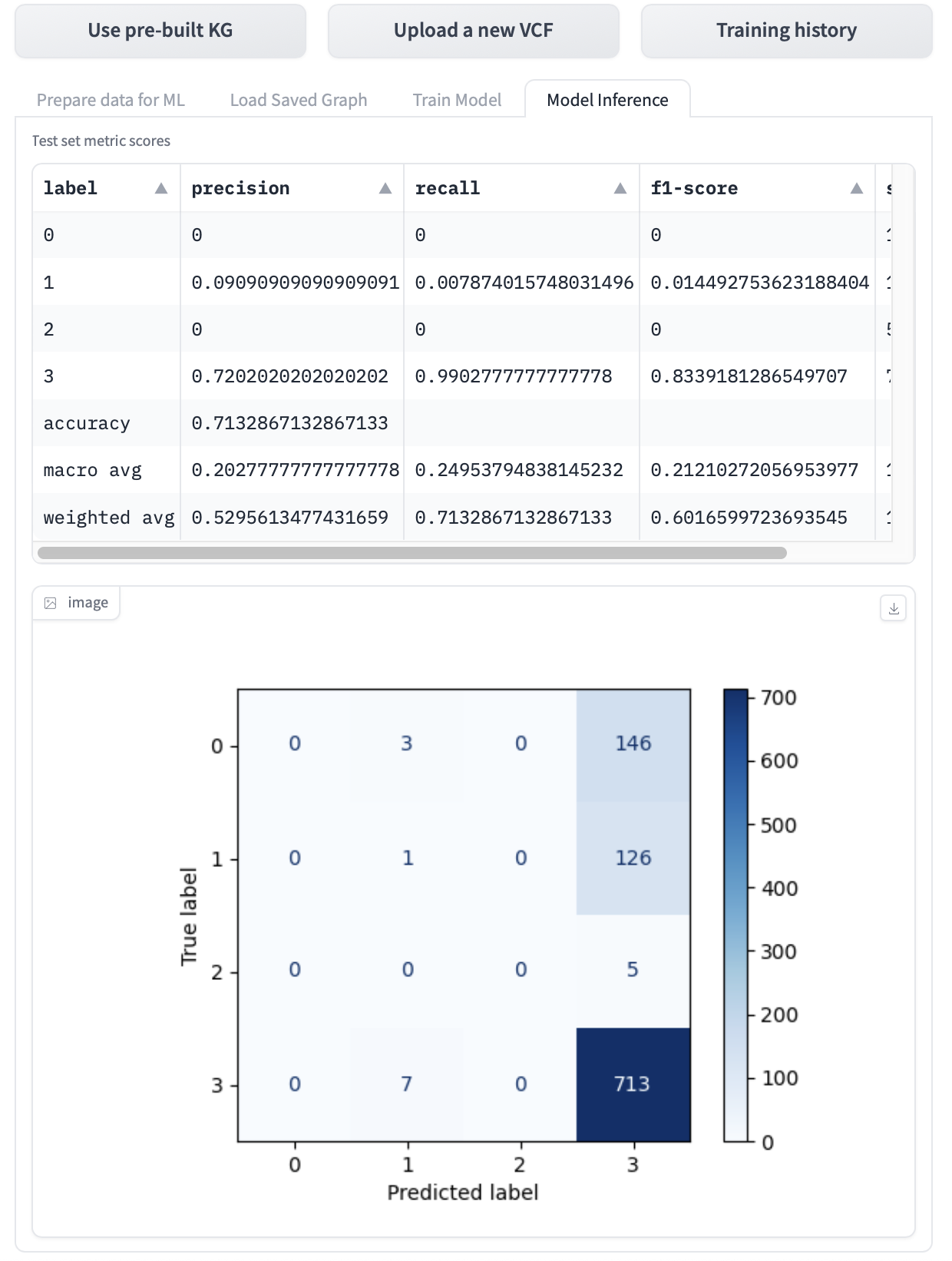}
    \caption{Graph ML Inferencing in VariantKG tool.}
    \label{fig:variant_kg_graph_inference}
\end{figure}

If the user wants to infer the ML task, they will be navigated to the `Inference' tab shown in Figure \ref{fig:variant_kg_graph_inference}, which displays the evaluation metrics and confusion matrix. The model is evaluated on Accuracy, macro-, and weighted- Precision, Recall, F1 score, and support, which is the number of samples for the given class. 

\section{Conclusion}
\label{conclusion}

This work shows that representing genomic data as knowledge graphs allows vast and diverse information to be integrated from various sources. Modeling entities as nodes and relationships as edges provides an ideal framework for integrating and organizing diverse information. We first described the data collection pipeline, followed by the usage of the SnpEff tool to obtain additional annotations and the SPARQLing Genomics tool to convert the annotations to an RDF format. An ontology to collate information gathered from different sources is presented. Using this ontology, we described how the knowledge graph is created. This knowledge graph contains RNA sequencing information at the variant level from COVID-19 patients from different regions such as lung, blood, etc. Lastly, we demonstrate the usage of the created knowledge graph for a node classification task using the Deep Graph Library through our tool, VariantKG. As part of this, we described various scenarios that can be performed by a user. As part of our future work, we aim to expand the knowledge graph and explore more avenues to use the same to aid researchers working in this domain.

\section{Acknowledgments}

This work was supported by the National Science Foundation under Grant Nos. 2201583 and 2034247.

\section{Data Availability Statement}
The raw VCF and CADD Scores used in the knowledge graph creation for this study can be found in the IEEE Dataport \cite{ieeedataport}. The tool can be accessed via VariantKG \footnote{\url{https://github.com/MU-Data-Science/GAF}}.









\bibliographystyle{unsrtnat}
\bibliography{references}  

\begin{thebibliography}{38}
\providecommand{\natexlab}[1]{#1}
\providecommand{\url}[1]{\texttt{#1}}
\expandafter\ifx\csname urlstyle\endcsname\relax
  \providecommand{\doi}[1]{doi: #1}\else
  \providecommand{\doi}{doi: \begingroup \urlstyle{rm}\Url}\fi

\bibitem[Olson(1993)]{olson1993human}
Maynard~V Olson.
\newblock The human genome project.
\newblock \emph{Proceedings of the National Academy of Sciences}, 90\penalty0 (10):\penalty0 4338--4344, 1993.

\bibitem[Rentzsch et~al.(2019)Rentzsch, Witten, Cooper, Shendure, and Kircher]{rentzsch2019cadd}
Philipp Rentzsch, Daniela Witten, Gregory~M Cooper, Jay Shendure, and Martin Kircher.
\newblock Cadd: predicting the deleteriousness of variants throughout the human genome.
\newblock \emph{Nucleic acids research}, 47\penalty0 (D1):\penalty0 D886--D894, 2019.

\bibitem[Rentzsch et~al.(2021)Rentzsch, Schubach, Shendure, and Kircher]{rentzsch2021cadd}
Philipp Rentzsch, Max Schubach, Jay Shendure, and Martin Kircher.
\newblock Cadd-splice—improving genome-wide variant effect prediction using deep learning-derived splice scores.
\newblock \emph{Genome medicine}, 13\penalty0 (1):\penalty0 1--12, 2021.

\bibitem[Rao et~al.(2021)Rao, Zachariah, Rao, Tonellato, Warren, and Simoes]{ieeedataport}
Praveen Rao, Arun Zachariah, Deepthi Rao, Peter Tonellato, Wesley Warren, and Eduardo Simoes.
\newblock Variant analysis of human genome sequences for covid-19 research.
\newblock 2021.
\newblock \doi{10.21227/b0ph-s175}.
\newblock URL \url{https://dx.doi.org/10.21227/b0ph-s175}.

\bibitem[Mahdisoltani et~al.(2013)Mahdisoltani, Biega, and Suchanek]{mahdisoltani2013yago3}
Farzaneh Mahdisoltani, Joanna Biega, and Fabian~M Suchanek.
\newblock Yago3: A knowledge base from multilingual wikipedias.
\newblock In \emph{CIDR}, 2013.

\bibitem[Vrande{\v{c}}i{\'c} and Kr{\"o}tzsch(2014)]{vrandevcic2014wikidata}
Denny Vrande{\v{c}}i{\'c} and Markus Kr{\"o}tzsch.
\newblock Wikidata: a free collaborative knowledgebase.
\newblock \emph{Communications of the ACM}, 57\penalty0 (10):\penalty0 78--85, 2014.

\bibitem[Lehmann et~al.(2015)Lehmann, Isele, Jakob, Jentzsch, Kontokostas, Mendes, Hellmann, Morsey, Van~Kleef, Auer, et~al.]{lehmann2015dbpedia}
Jens Lehmann, Robert Isele, Max Jakob, Anja Jentzsch, Dimitris Kontokostas, Pablo~N Mendes, Sebastian Hellmann, Mohamed Morsey, Patrick Van~Kleef, S{\"o}ren Auer, et~al.
\newblock Dbpedia--a large-scale, multilingual knowledge base extracted from wikipedia.
\newblock \emph{Semantic web}, 6\penalty0 (2):\penalty0 167--195, 2015.

\bibitem[Guha et~al.(2016)Guha, Brickley, and Macbeth]{guha2016schema}
Ramanathan~V Guha, Dan Brickley, and Steve Macbeth.
\newblock Schema. org: evolution of structured data on the web.
\newblock \emph{Communications of the ACM}, 59\penalty0 (2):\penalty0 44--51, 2016.

\bibitem[Dong(2018)]{dong2018challenges}
Xin~Luna Dong.
\newblock Challenges and innovations in building a product knowledge graph.
\newblock In \emph{Proceedings of the 24th ACM SIGKDD International conference on knowledge discovery \& data mining}, pages 2869--2869, 2018.

\bibitem[Gharibi et~al.(2020)Gharibi, Zachariah, and Rao]{gharibi2020foodkg}
Mohamed Gharibi, Arun Zachariah, and Praveen Rao.
\newblock Foodkg: A tool to enrich knowledge graphs using machine learning techniques.
\newblock \emph{Frontiers in big Data}, 3:\penalty0 12, 2020.

\bibitem[Feng et~al.(2024)Feng, Tang, Gao, Zhu, Li, Yang, Yao, Huang, and Liu]{feng2024genomickb}
Fan Feng, Feitong Tang, Yijia Gao, Dongyu Zhu, Tianjun Li, Shuyuan Yang, Yuan Yao, Yuanhao Huang, and Jie Liu.
\newblock 1999-lb: Genomickb—a knowledge graph for human genomic data to advance understanding of diabetes.
\newblock \emph{Diabetes}, 73\penalty0 (Supplement\_1):\penalty0 1999--LB, June 2024.
\newblock \doi{10.2337/db24-1999-LB}.
\newblock URL \url{https://doi.org/10.2337/db24-1999-LB}.

\bibitem[Feng et~al.(2023)Feng, Tang, Gao, Zhu, Li, Yang, Yao, Huang, and Liu]{feng2023genomickb}
Fan Feng, Feitong Tang, Yijia Gao, Dongyu Zhu, Tianjun Li, Shuyuan Yang, Yuan Yao, Yuanhao Huang, and Jie Liu.
\newblock Genomickb: a knowledge graph for the human genome.
\newblock \emph{Nucleic Acids Research}, 51\penalty0 (D1):\penalty0 D950--D956, 2023.

\bibitem[Sakor et~al.(2023)Sakor, Gr{\"o}nli, M{\o}rland, Ruiz, Mikalsen, and Ghosh]{sakor2023knowledge4covid19}
Ahmad Sakor, Tor-Morten Gr{\"o}nli, Berit M{\o}rland, Elisa Ruiz, Marte Mikalsen, and Subrata Ghosh.
\newblock Knowledge4covid-19: A semantic-based approach for constructing a covid-19 related knowledge graph from various sources and analyzing treatments’ toxicities.
\newblock \emph{Journal of Web Semantics}, 75:\penalty0 100760, Jan 2023.
\newblock \doi{10.1016/j.websem.2022.100760}.
\newblock URL \url{https://doi.org/10.1016/j.websem.2022.100760}.

\bibitem[Reese et~al.(2021)Reese, Unni, Callahan, Cappelletti, Ravanmehr, Carbon, Shefchek, Good, Balhoff, Fontana, et~al.]{reese2021kg}
Justin~T Reese, Deepak Unni, Tiffany~J Callahan, Luca Cappelletti, Vida Ravanmehr, Seth Carbon, Kent~A Shefchek, Benjamin~M Good, James~P Balhoff, Tommaso Fontana, et~al.
\newblock Kg-covid-19: a framework to produce customized knowledge graphs for covid-19 response.
\newblock \emph{Patterns}, 2\penalty0 (1), 2021.

\bibitem[Chen et~al.(2021)Chen, Zhang, Chiam, Zhou, Lee, Tan, Koh, Li, and Zhang]{chen2021covid19}
Chuming Chen, Xueying Zhang, Yuanhao Chiam, Mingyang Zhou, Veronica~J. Lee, Thanh~G. Tan, Phillip~W. Koh, Haixu Li, and Zhiyong Zhang.
\newblock Covid-19 knowledge graph from semantic integration of biomedical literature and databases.
\newblock \emph{Bioinformatics}, 37\penalty0 (23):\penalty0 4597--4598, Dec 2021.
\newblock \doi{10.1093/bioinformatics/btab694}.
\newblock URL \url{https://doi.org/10.1093/bioinformatics/btab694}.

\bibitem[Liu et~al.(2020)Liu, Hu, Jiang, and Zhou]{liu2020deepcdr}
Qiao Liu, Zhiqiang Hu, Rui Jiang, and Mu~Zhou.
\newblock Deepcdr: a hybrid graph convolutional network for predicting cancer drug response.
\newblock \emph{Bioinformatics}, 36\penalty0 (Supplement\_2):\penalty0 i911--i918, 2020.
\newblock \doi{https://doi.org/10.1093/bioinformatics/btaa822}.

\bibitem[Lanchantin and Qi(2019)]{lanchantin2019graph}
Jack Lanchantin and Yanjun Qi.
\newblock Graph convolutional networks for epigenetic state prediction using both sequence and 3d genome data.
\newblock \emph{BioRxiv}, page 840173, 2019.
\newblock \doi{https://doi.org/10.1101/840173}.

\bibitem[Harnoune et~al.(2021)Harnoune, Rhanoui, Mikram, Yousfi, Elkaimbillah, and El~Asri]{harnoune2021bertclinical}
Ayoub Harnoune, Maryem Rhanoui, Mounia Mikram, Siham Yousfi, Zineb Elkaimbillah, and Bouchra El~Asri.
\newblock Bert based clinical knowledge extraction for biomedical knowledge graph construction and analysis.
\newblock \emph{Computer Methods and Programs in Biomedicine Update}, 1:\penalty0 100042, 2021.
\newblock \doi{https://doi.org/10.1016/j.cmpbup.2021.100042}.

\bibitem[Domingo-Fern{\'a}ndez et~al.(2021)Domingo-Fern{\'a}ndez, Baksi, Schultz, Gadiya, Karki, Raschka, Ebeling, and Hofmann-Apitius]{domingo-fernandez2021covid19}
Daniel Domingo-Fern{\'a}ndez, Swayamsiddha Baksi, Bastien Schultz, Yash Gadiya, Rajendra Karki, Tjark Raschka, Claas Ebeling, and Martin Hofmann-Apitius.
\newblock Covid-19 knowledge graph: A computable, multi-modal, cause-and-effect knowledge model of covid-19 pathophysiology.
\newblock \emph{Bioinformatics}, 37\penalty0 (9):\penalty0 1332--1334, June 2021.
\newblock \doi{10.1093/bioinformatics/btaa834}.
\newblock URL \url{https://doi.org/10.1093/bioinformatics/btaa834}.

\bibitem[Al-Obeidat et~al.(2020)Al-Obeidat, Ballout, Shehab, and Awajan]{al-obeidat2020cone-kg}
Feras Al-Obeidat, Mark Ballout, Muhammad Shehab, and Anoud Awajan.
\newblock Cone-kg: A semantic knowledge graph with news content and social context for studying covid-19 news articles on social media.
\newblock In \emph{2020 Seventh International Conference on Social Networks Analysis, Management and Security (SNAMS)}, pages 1--7. IEEE, 2020.
\newblock \doi{10.1109/SNAMS52053.2020.9336541}.
\newblock URL \url{https://doi.org/10.1109/SNAMS52053.2020.9336541}.

\bibitem[Sun et~al.(2018)Sun, Chang, Zhao, and Long]{sun2018knowledge}
Wenli Sun, Changgee Chang, Yize Zhao, and Qi~Long.
\newblock Knowledge-guided bayesian support vector machine for high-dimensional data with application to analysis of genomics data.
\newblock In \emph{2018 IEEE International Conference on Big Data (Big Data)}, pages 1484--1493. IEEE, 2018.

\bibitem[Li et~al.(2008)Li, Ruan, and Durbin]{li2008mapping}
Heng Li, Jue Ruan, and Richard Durbin.
\newblock Mapping short dna sequencing reads and calling variants using mapping quality scores.
\newblock \emph{Genome research}, 18\penalty0 (11):\penalty0 1851--1858, 2008.

\bibitem[Li and Durbin(2009)]{li2009fast}
Heng Li and Richard Durbin.
\newblock Fast and accurate short read alignment with burrows--wheeler transform.
\newblock \emph{bioinformatics}, 25\penalty0 (14):\penalty0 1754--1760, 2009.

\bibitem[McKenna et~al.(2010)McKenna, Hanna, Banks, Sivachenko, Cibulskis, Kernytsky, Garimella, Altshuler, Gabriel, Daly, et~al.]{mckenna2010genome}
Aaron McKenna, Matthew Hanna, Eric Banks, Andrey Sivachenko, Kristian Cibulskis, Andrew Kernytsky, Kiran Garimella, David Altshuler, Stacey Gabriel, Mark Daly, et~al.
\newblock The genome analysis toolkit: a mapreduce framework for analyzing next-generation dna sequencing data.
\newblock \emph{Genome research}, 20\penalty0 (9):\penalty0 1297--1303, 2010.

\bibitem[Danecek et~al.(2011)Danecek, Auton, Abecasis, Albers, Banks, DePristo, Handsaker, Lunter, Marth, Sherry, et~al.]{danecek2011variant}
Petr Danecek, Adam Auton, Goncalo Abecasis, Cornelis~A Albers, Eric Banks, Mark~A DePristo, Robert~E Handsaker, Gerton Lunter, Gabor~T Marth, Stephen~T Sherry, et~al.
\newblock The variant call format and vcftools.
\newblock \emph{Bioinformatics}, 27\penalty0 (15):\penalty0 2156--2158, 2011.

\bibitem[Cingolani et~al.(2012)Cingolani, Platts, Coon, Nguyen, Wang, Land, Lu, and Ruden]{cingolani2012program}
P.~Cingolani, A.~Platts, M.~Coon, T.~Nguyen, L.~Wang, S.J. Land, X.~Lu, and D.M. Ruden.
\newblock A program for annotating and predicting the effects of single nucleotide polymorphisms, snpeff: Snps in the genome of drosophila melanogaster strain w1118; iso-2; iso-3.
\newblock \emph{Fly}, 6\penalty0 (2):\penalty0 80--92, 2012.

\bibitem[Di~Bartolomeo et~al.(2018)Di~Bartolomeo, Pepe, Savo, and Santarelli]{di2018sparqling}
Sara Di~Bartolomeo, Gianluca Pepe, Domenico~Fabio Savo, and Valerio Santarelli.
\newblock Sparqling: Painlessly drawing sparql queries over graphol ontologies.
\newblock In \emph{VOILA@ ISWC}, pages 64--69, 2018.
\newblock URL \url{https://github.com/UMCUGenetics/sparqling-genomics/tree/main?tab=readme-ov-file}.

\bibitem[Bolleman et~al.(2016)Bolleman, Mungall, Strozzi, Baran, Dumontier, Bonnal, Buels, Hoehndorf, Fujisawa, Katayama, et~al.]{bolleman2016faldo}
Jerven~T Bolleman, Christopher~J Mungall, Francesco Strozzi, Joachim Baran, Michel Dumontier, Raoul~JP Bonnal, Robert Buels, Robert Hoehndorf, Takatomo Fujisawa, Toshiaki Katayama, et~al.
\newblock Faldo: a semantic standard for describing the location of nucleotide and protein feature annotation.
\newblock \emph{Journal of biomedical semantics}, 7\penalty0 (1):\penalty0 1--12, 2016.

\bibitem[bla(2024)]{blazegraph}
Blazegraph db.
\newblock \url{https://blazegraph.com}, 2024.

\bibitem[Katib et~al.(2017)Katib, Rao, and Slavov]{RIQ2017}
Anas Katib, Praveen Rao, and Vasil Slavov.
\newblock {A Tool for Efficiently Processing SPARQL Queries on RDF Quads}.
\newblock In \emph{International Semantic Web Conference (Posters, Demos \& Industry Tracks)}, 2017.

\bibitem[Katib et~al.(2016)Katib, Slavov, and Rao]{RIQ2016}
Anas Katib, Vasil Slavov, and Praveen Rao.
\newblock {RIQ: Fast processing of SPARQL queries on RDF quadruples}.
\newblock \emph{Journal of Web Semantics}, 37-38:\penalty0 90--111, 2016.
\newblock ISSN 1570-8268.

\bibitem[Slavov et~al.(2015)Slavov, Katib, Rao, Paturi, and Barenkala]{RIQ2015}
Vasil Slavov, Anas Katib, Praveen Rao, Srivenu Paturi, and Dinesh Barenkala.
\newblock {Fast processing of SPARQL queries on RDF quadruples}.
\newblock \emph{arXiv preprint arXiv:1506.01333}, 2015.

\bibitem[Wang et~al.(2019)Wang, Zheng, Ye, Gan, Li, Song, Zhou, Ma, Yu, Gai, Xiao, He, Karypis, Li, and Zhang]{wang2019dgl}
Minjie Wang, Da~Zheng, Zihao Ye, Quan Gan, Mufei Li, Xiang Song, Jinjing Zhou, Chao Ma, Lingfan Yu, Yu~Gai, Tianjun Xiao, Tong He, George Karypis, Jinyang Li, and Zheng Zhang.
\newblock Deep graph library: A graph-centric, highly-performant package for graph neural networks.
\newblock \emph{arXiv preprint arXiv:1909.01315}, 2019.

\bibitem[Zhang et~al.(2019)Zhang, Tong, Xu, and Maciejewski]{zhang2019graph}
Si~Zhang, Hanghang Tong, Jiejun Xu, and Ross Maciejewski.
\newblock Graph convolutional networks: a comprehensive review.
\newblock \emph{Computational Social Networks}, 6\penalty0 (1):\penalty0 1--23, 2019.

\bibitem[Hamilton et~al.(2017)Hamilton, Ying, and Leskovec]{hamilton2017sage}
Will Hamilton, Zhitao Ying, and Jure Leskovec.
\newblock Inductive representation learning on large graphs.
\newblock In \emph{Advances in neural information processing systems}, page 30.d, 2017.

\bibitem[Kingma and Ba(2014)]{kingma2014adam}
Diederik~P Kingma and Jimmy Ba.
\newblock Adam: A method for stochastic optimization.
\newblock \emph{arXiv preprint arXiv:1412.6980}, 2014.

\bibitem[Ricci et~al.(2014)Ricci, Eide, and Team]{ricci2014introducing}
Robert Ricci, Eric Eide, and CloudLab Team.
\newblock Introducing cloudlab: Scientific infrastructure for advancing cloud architectures and applications.
\newblock \emph{; login:: the magazine of USENIX \& SAGE}, 39\penalty0 (6):\penalty0 36--38, 2014.

\bibitem[Abid et~al.(2019)Abid, Abdalla, Abid, Khan, Alfozan, and Zou]{abid2019gradio}
Abubakar Abid, Ali Abdalla, Ali Abid, Dawood Khan, Abdulrahman Alfozan, and James Zou.
\newblock Gradio: Hassle-free sharing and testing of ml models in the wild.
\newblock \emph{arXiv preprint arXiv:1906.02569}, 2019.

\end{thebibliography}






\end{document}